\pdfoutput=1

\documentclass{article} %
\usepackage{iclr2027_conference,times}

\usepackage{amsmath,amsfonts,bm}

\def\eqref#1{equation~\ref{#1}}

\def\1{\bm{1}}

\def\vtheta{{\bm{\theta}}}

\DeclareMathAlphabet{\mathsfit}{\encodingdefault}{\sfdefault}{m}{sl}
\SetMathAlphabet{\mathsfit}{bold}{\encodingdefault}{\sfdefault}{bx}{n}

\usepackage{hyperref}
\usepackage{url}
\usepackage{booktabs}
\usepackage{subcaption}
\usepackage{multirow}
\usepackage{graphicx}
\usepackage{wrapfig} %
\usepackage{subcaption}
\usepackage[table]{xcolor}
\usepackage{array}
\usepackage{soul}
\usepackage{float}
\newcommand{\modelName}{\text{Argos}}

\newif\ifshowcomments
\showcommentstrue %

\title{\modelName{}: Adapt Rich Geometric Priors for Generalizable Online Scene-Change-Detection}

\author{Ruihan Xu\textsuperscript{\normalfont 1} \quad
Jiae Yoon\textsuperscript{\normalfont 2} \quad
Kaichen Zhou\textsuperscript{\normalfont 1} \quad
Ue-Hwan Kim\textsuperscript{\normalfont 2} \quad
Luca Carlone\textsuperscript{\normalfont 1} \\
{\normalfont \textsuperscript{1}MIT \quad
\textsuperscript{2}Gwangju Institute of Science and Technology} \\
{\normalfont\small \{multyxu, zhouk777, lcarlone\}@mit.edu \quad
jiaeyoon@gm.gist.ac.kr \quad
uehwan@gist.ac.kr}
}

\iclrfinalcopy %
\begin{document}
\newcommand{\datasetComparison}{
\begin{table*}[t]
    \centering
    \tiny
    \caption{\textbf{Dataset comparison.} Places$^{*}$ and single-visit video length (seconds) are
    reported as mean\,/\,min\,/\,max. Pairs counts cross-visit comparisons and Views counts frames
    with change masks. $^*$Approximated through appearance clustering.}
    \label{tab:dataset}
    \begin{tabular}{llcccccccc}
        \toprule
        \multicolumn{2}{c}{Dataset} & Pairs & Views & Scenes & Places$^{*}$ & Length\,(s) & Modality & Viewpoint & Trajectory \\
        \midrule
        \multirow{5}[1]{4.5em}{Released datasets} & VL-CMU-CD & 1{,}362 & 1{,}362 & 152 & -- & -- & Real & Identical & - \\
        &PSCD & 770 & 1{,}540 & -- & -- & -- & Real & Identical & - \\
        &ChangeSim & 20 & 130\,K & 10 & -- & -- & Sim. & Similar &  Manual \\
        \cmidrule(lr){2-10}
        &VSCD & 1{,}041 & 1.13\,M & 210 & 1.6/1/7 & 36.0/10.1/116.0 & both & Free & Manual\\
        &SceneDiff Dataset & 355 & 172\,K & 355 & 1.1/1/2 & 8.1/2.4/26.1 & Real & Free & Manual \\
        \midrule
        \multirow{3}[1]{4.5em}{\textbf{\modelName{}-CD \\(Ours)}} & \modelName{}-CD-HSSD & 501 & 644\,K & 167 & \textbf{3.8/1/11} & \textbf{ 42.8/4.0/203.6 } & Sim. & Free & Automatic \\
        &\modelName{}-CD-SceneSmith & 612 & 138\,K & 204 & 1.5/1/5 & 7.5/1.2/51.7 & Sim. & Free & Automatic \\
        &\modelName{}-CD-real & 19 & 12{,}367 & 19 & 1.0/1/1 & 10.8/4.8/23.1 & Real & Free & Manual \\
        \bottomrule
    \end{tabular}%
\end{table*}
}

\newcommand{\mainTable}{
\begin{table}[t]
    \centering
    \tiny
    \caption{\textbf{Real-world generalization study.} \modelName{} performs consistently better than learned baselines even without fine-tuning to the real-world datasets. $^\dagger$ indicates zero-shot methods.  }
    \label{tab:main_table}
    \begin{tabular}{clcccccccc}
        \toprule
        \multicolumn{2}{c}{\multirow{2}[1]{*}{\textbf{Method}}} & \multicolumn{4}{c}{SceneDiff Dataset} & \multicolumn{4}{c}{\modelName{}-CD-real} \\
        \cmidrule(lr){3-6}\cmidrule(lr){7-10}
         && Static & Change & mIoU & F1 & Static & Change & mIoU & F1 \\
        \midrule
        \multirow{6}[1]{*}{SCD}&CSCDNet-finetune & 93.33 & 15.25 & 54.29 & 20.23 & 92.61 & 10.64 & 51.63 & 13.60\\
        &C-3PO$^*$-finetune     & 94.42 & 17.88 & 56.15 & 22.47 & 92.26 & 6.96 & 49.61 & 10.86 \\
        &GeSCF$^\dagger$ & 69.69 & 5.04 & 37.37 & 15.12 & 63.61 & 6.15 & 34.88 & 13.50 \\
        &RobustSCD-finetune  & 94.41 & 18.97 & 56.69 & 21.25 & 91.78 & 13.50 & 52.64 & 16.55 \\
        &ZSSCD$^\dagger$ & 88.63 & 6.74 & 47.69 & 13.22 & 77.53 & 3.12 & 40.33 & 8.27 \\
        &TERDNet-finetune & 94.30 & 20.32 & 57.30 & 25.14 & 91.78 & 11.38 & 51.58 & 16.56 \\
        \midrule
        \multirow{7}[1]{*}{VSCD}&VSCDNet & 91.40 & 10.48 & 50.94 & 15.46 & 87.22 & 9.05 & 48.14 & 13.39 \\
        &VSCDNet-finetune & 95.72 & 20.70 & 58.21 & 23.84 & 93.71 & 8.97 & 51.34 & 14.61 \\
        &VSCDNet-foundation & 95.85 & 19.27 & 57.56 & 24.26 & 90.76 & 8.68 & 49.72 & 16.51 \\
        &SceneDiff$^\dagger$ & 96.48 & 30.31 & 63.40 & 37.58 & 95.55 & 32.92 & 64.23 & 35.52 \\
        \cmidrule(lr){2-10}
        &\textbf{Ours-synthetic-only} & 97.13 & 39.84 & 68.48 & 37.42 & 94.74 & 31.18 & 62.96 & 33.95 \\
        &\textbf{Ours-finetune} & \underline{98.62} & \underline{59.91} & \underline{79.27} & \underline{44.73} & \textbf{97.13} & \textbf{50.92} & \textbf{74.03} & \textbf{41.51} \\
        &\textbf{Ours-foundation} & \textbf{98.78} & \textbf{62.54} & \textbf{80.66} & \textbf{45.98} & \underline{97.08} & \underline{47.74} & \underline{72.41} & \underline{41.24} \\
        \bottomrule
    \end{tabular}
\end{table}
}

\newcommand{\customCDTable}{
\begin{table}[t]
    \centering
    \tiny
    \caption{\textbf{Performance on Video Scene Change Detection.} $^*$C-3PO architecture disables the ``remove'' feature branch. $^{**}$The VSCD dataset is modified to contain ``appeared'' labels in both reference and query frame.}
    \label{tab:custom_cd}
    \begin{tabular}{lcccccccccccc}
        \toprule
        \multirow{2}[1]{*}{\textbf{Method}} & \multicolumn{4}{c}{\modelName{}-CD-HSSD} &  \multicolumn{4}{c}{\modelName{}-CD-SceneSmith} & \multicolumn{4}{c}{VSCD$^{**}$} \\
        \cmidrule(lr){2-5}\cmidrule(lr){6-9}\cmidrule(lr){10-13}
        & Static & Change & mIoU & F1 & Static & Change & mIoU & F1 & Static & Change & mIoU & F1 \\
        \midrule
        CSCDNet   & 92.91 & 38.10 & 65.51 & 36.53 & 94.32 & 31.11 & 62.72 & 34.83 & 94.03 & 31.95 & 62.99 & 32.44\\
        C-3PO$^*$  & 91.85 & 37.65 & 64.75 & 37.68 & 94.46 & 38.68 & 66.57 & 40.77 & 93.88 & 32.98 & 63.43 & 32.90 \\
        GeSCF$^\dagger$     & 76.73 & 13.81 & 45.27 & 28.41 & 81.56 & 13.72 & 47.64 & 29.80 & 77.74 & 12.26 & 45.00 & 23.98 \\
        RobustSCD & 88.86 & 33.18 & 61.02 & 34.76 & 92.70 & 33.92 & 63.30 & 36.85 & 89.54 & 27.18 & 58.36 & 30.59 \\
        ZSSCD$^\dagger$     & 90.03 & 16.58 & 53.31 & 16.22 & 91.63 & 15.15 & 53.39 & 15.49 & 93.00 & 15.72 & 54.36 & 16.02 \\
        TERDNet   & 93.05 & 42.14 & 67.59 & 40.62 & 95.19 & 38.58 & 66.88 & 40.31 & 90.90 & 27.92 & 59.41 & 30.92 \\
        \midrule
        VSCDNet & 93.50 & 42.71 & 68.11 & 41.77 & 94.62 & 38.94 & 66.78 & 41.17 & 94.49 & 34.55 & 64.52 & 36.31 \\
        SceneDiff$^\dagger$ & 93.80 & 33.44 & 63.62 & 32.44  & 96.38 & 40.42 & 68.40 & 36.58 & 96.71 & 48.21 & 72.46 & 38.70 \\
        \midrule
        \textbf{Ours} & \textbf{98.38} & \textbf{78.64} & \textbf{88.51} & \textbf{62.16} & \textbf{99.08} & \textbf{82.43} & \textbf{90.75} & \textbf{69.08} & \textbf{98.24} & \textbf{68.69} & \textbf{83.46} & \underline{53.14} \\
        \textbf{Ours-foundation} & \underline{98.07} & \underline{75.51} & \underline{86.79} & \underline{60.69} & \underline{98.91} & \underline{80.41} & \underline{89.66} & \underline{68.93} & \underline{98.05} & \underline{68.52} & \underline{83.28} & \textbf{53.33} \\
        \bottomrule
    \end{tabular}
\end{table}
}

\newcommand{\scdTable}{
\begin{table}[t]
    \centering
    \tiny
    \caption{\textbf{Performance on paired-image Scene Change Detection.} \modelName{} outperforms all baseline methods, and the foundation variant of \modelName{} scores closely to the dataset specific variants. }
    \label{tab:scd}
    \begin{tabular}{lcccccccccccc}
        \toprule
        \multirow{2}[1]{*}{\textbf{Method}} & \multicolumn{4}{c}{PSCD} & \multicolumn{4}{c}{ChangeSim$^*$} & \multicolumn{4}{c}{VL-CMU-CD} \\
        \cmidrule(lr){2-5}\cmidrule(lr){6-9}\cmidrule(lr){10-13}
        & Static & Change & mIoU & F1 & Static & Change & mIoU & F1 & Static & Change & mIoU & F1 \\
        \midrule
        FC-EF        & 84.97 & 37.99 & 61.48 & 48.14 & 66.21 & 12.44 & 39.32 & 20.72 & 91.36 & 36.72 & 64.04 & 45.61 \\
        FC-Siam-diff & 89.55 & 43.72 & 66.64 & 52.67 & 85.77 & 21.12 & 53.44 & 33.61 & 96.17 & 55.97 & 76.07 & 64.21 \\
        FC-Siam-conc & 86.96 & 41.76 & 64.36 & 52.13 & 81.63 & 19.01 & 50.32 & 29.40 & 94.40 & 50.81 & 72.61 & 58.71* \\
        CSCDNet      & 92.57 & 48.84 & 70.70 & 60.81 & 92.77 & 26.48 & 59.62 & 37.64 & 97.57 & 69.17 & 83.37 & 76.90 \\
        DR-TANet     & 91.17 & 43.40 & 67.28 & 55.79 & 90.91 & 27.57 & 59.24 & 38.53 & 97.16 & 65.81 & 81.48 & 73.98 \\
        C-3PO$^*$        & 92.36 & 50.19 & 71.27 & 60.66 & 93.00 & 28.68 & 60.84 & 38.92 & 97.86 & 72.33 & 85.10 & 80.00 \\
        ZSSCD$^\dagger$        & 83.41 & 19.94 & 51.67 & 28.94 & 94.55 & 30.03 & 62.29 & 40.51 & 93.12 & 36.14 & 64.63 & 50.14 \\
        RobustSCD    & 81.69 & 35.05 & 58.37 & 44.19 & 88.62 & 26.39 & 57.50 & 38.34 & 97.58 & 70.48 & 84.28 & 79.44 \\
        GeSCF$^\dagger$        & 81.63 & 26.16 & 53.89 & 41.76 & 90.39 & 16.23 & 53.31 & 27.04 & 96.63 & 58.81 & 77.72 & 75.39 \\
        TERDNet      & 93.56 & 55.30 & 74.43 & 65.81 & 93.17 & 35.07 & 64.12 & 46.24 & \underline{98.02} & \underline{73.80} & \underline{85.91} & \underline{83.43} \\
        \midrule
        \textbf{Ours} & \textbf{94.61} & \textbf{62.83} & \textbf{78.72} & \textbf{69.65} & \underline{95.02} & \textbf{50.03} & \textbf{72.52} & \textbf{61.73} & \textbf{98.28} & \textbf{77.80} & \textbf{88.03} & \textbf{86.3} \\
        \textbf{Ours-foundation} & \underline{93.69} & \underline{58.90} & \underline{76.34} & \underline{66.92} & \textbf{95.56} & \underline{46.33} & \underline{70.95} & \underline{54.97} & 97.92 & 73.33 & 85.63 & 81.61\\
        \bottomrule
    \end{tabular}
\end{table}
}

\newcommand{\ArchitectureAblationTable}{

\begin{table}[!thb]
    \centering
    \tiny
    \caption{\textbf{Architecture Ablation.} Inter-session cross attention and Camera reader complement each other to achieve the best performance, while VGGT-$\Omega$  backbone provides a strong foundation. All backbones are frozen; DINOv2 is a 2D self-supervised model, while $\pi^3$ and VGGT-$\Omega$ are geometric foundation models. }
    \label{tab:architecture_ablation}
    \begin{tabular}{cccccccccccc}
        \toprule
        \multicolumn{4}{c}{\textbf{Architecture Variant}} & \multicolumn{4}{c}{VSCD**} & \multicolumn{4}{c}{PSCD} \\
        \cmidrule(lr){0-3} \cmidrule(lr){5-8} \cmidrule(lr){9-12}
        Backbone & DPTHead & CrossAttention & CameraReader & Static & Change & mIoU & F1 & Static & Change & mIoU & F1 \\
        \midrule
        SAM & $\checkmark$ & $\checkmark$ & $\checkmark$           & 91.92 & 27.89 & 59.90 & 30.64 & 92.05 & 46.08 & 69.07 & 58.59\\
        DINOv2 & $\checkmark$ & $\checkmark$ & $\checkmark$        & 96.83 & 50.07 & 73.45 & 41.32 & 93.39 & 55.29 & 74.34 & 65.37\\
        $\pi^3$ & $\checkmark$ & $\checkmark$ & $\checkmark$       & 97.98 & 64.39 & 81.18 & 50.83 & 93.13 & 55.11 & 74.12 & 64.44\\
        \midrule
        VGGT-$\Omega$ & $\checkmark$ & $\times$ & $\times$         & 96.34 & 47.97 & 72.16 & 42.52 & \underline{94.64} & \underline{63.88} & \underline{79.27} & \underline{69.26}  \\
        VGGT-$\Omega$ & $\checkmark$ & $\checkmark$ & $\times$     & \underline{97.89} & \underline{63.51} & \underline{80.70} & \underline{50.01} & 94.48 & 62.11 & 78.29 & 69.08   \\
        VGGT-$\Omega$ & $\checkmark$ & $\times$ & $\checkmark$     & 93.74 & 25.96 & 59.85 & 29.37 & 94.63 & 63.53 & 79.08 & 69.16 \\
        VGGT-$\Omega$ & $\checkmark$ & $\checkmark$ & $\checkmark$ & \textbf{98.24} & \textbf{68.69} & \textbf{83.46} & \textbf{53.14} & \textbf{94.80} & \textbf{64.56} & \textbf{79.68} & \textbf{69.89}\\
        \bottomrule
    \end{tabular}
\end{table}
}

\newcommand{\oscdPrescreenTable}{
\begin{table}[tbh]
    \centering
    \scriptsize
    \caption{\textbf{O-SCD pose-registration prescreen coverage.} Across all 220 classic-VSCD test candidates, increasing the reference-frame budget from 32 to 128 provide enough datapoint for reliable evaluation.}
    \label{tab:oscd_prescreen}
    \begin{tabular}{lccccc}
        \toprule
        Setting & \#Samples & COLMAP Success & Avg.\ Query Frames & Full Success (32/32) & $\geq$90\% Success (29+/32) \\
        \midrule
        32+32  & 220 & 141 (64.1\%) & 7.5 / 32  & 4 (1.8\%) & 4 (1.8\%) \\
        64+32  & 220 & 181 (82.3\%) & 11.9 / 32 & 6 (2.7\%) & 15 (6.8\%) \\
        128+32 & 220 & 199 (90.5\%) & 13.8 / 32 & 7 (3.2\%) & 15 (6.8\%) \\
        256+32 & 220 & 195 (88.6\%) & 13.0 / 32	4 & (1.8\%) & 9 (4.1\%) \\
        \bottomrule
    \end{tabular}
\end{table}
}

\newcommand{\oscdResultTable}{
\begin{table}[tbh]
    \centering
    \scriptsize
    \caption{\textbf{Comparison against O-SCD on the top-12 classic-VSCD scenes O-SCD can register.} When given enough reference frames, O-SCD outperforms the VSCDNet baseline but still falls short of our proposed method.}
    \label{tab:oscd_result}
    \begin{tabular}{lcccc}
        \toprule
        Method & Static & Change & mIoU & F1 \\
        \midrule
        O-SCD (R=32)  & 88.42 & 19.18 & 53.80 & 26.28 \\
        O-SCD (R=128) & 90.20 & \underline{28.95} & \underline{59.57} & \underline{40.64} \\
        VSCDNet       & \underline{90.22} & 22.45 & 56.33 & 31.05 \\
        Ours (VSCD)   & \textbf{93.21} & \textbf{37.70} & \textbf{65.46} & \textbf{49.41} \\
        \bottomrule
    \end{tabular}
\end{table}
}

\newcommand{\vscdBaselineTable}{
\definecolor{lightred}{RGB}{255,220,220}
\begin{table}[tbh]
    \fontsize{5.4pt}{6.4pt}\selectfont %
    \centering
    \caption{
    \textbf{Overall performance on the original VSCD benchmark.}
    We report frame-wise F1 score (\%) on the synthetic dataset and the real-world test set, together with results stratified by video length, graphic level, and the number of changed objects.
    }
    \label{tab:original_vscd}
    \begin{tabular}{lc|ccc|cc|ccc|>{\columncolor{lightred}}c|cc|>{\columncolor{lightred}}c}
        \toprule
        \multicolumn{2}{c|}{Method} & \multicolumn{3}{c|}{Video length} & \multicolumn{2}{c|}{Graphic level} & \multicolumn{3}{c|}{Object changes} & Synthetic & \multicolumn{2}{c|}{Captured by} & Real-world \\
        Task & Model & Low & Mid & High & Mid & High & Low & Mid & High & Overall & Robot & Human & Overall\\
        \midrule
        \multirow{2}[1]{*}{AOD} & TCF-LMO  & 17.1 & 20.1 & 22.2 & 18.7 & 20.8 & 15.2 & 21.0 & 21.8 & 19.7 & 7.4 & 12.0 & 10.3 \\
        & PBCD-MC  & 24.6 & 28.0 & 26.5 & 26.3 & 27.4 & 23.4 & 28.0 & 27.8 & 26.8 & 16.8 & 15.7 & 16.1 \\
        \midrule
        \multirow{5}[1]{*}{SCD} & CSCDNet  & 18.8 & 21.4 & 16.9 & 21.7 & 17.5 & 20.0 & 20.1 & 19.0 & 19.8 & 8.2 & 9.6 & 9.1 \\
        & DR-TANet & 20.5 & 22.0 & 17.2 & 22.9 & 17.9 & 20.8 & 20.5 & 20.6 & 20.6 & 9.1 & 13.1 & 11.6 \\
        & C-3PO & 22.7 & 26.1 & 20.8 & 30.1 & 16.9 & 18.8 & 25.1 & 27.8 & 24.1 & 8.0 & 13.9 & 11.7 \\
        & ZSSCD & 0.1 & 0.5 & 0.1 & 0.5 & 0.1 & 0.2 & 0.4 & 0.2 & 0.3 & 2.1 & 0.0 & 0.8 \\
        & GeSCF & 26.4 & 30.4 & 31.2 & 29.3 & 29.8 & 28.0 & 29.7 & 30.6 & 29.5 & 25.1 & 12.6 & 17.3 \\
        \midrule
        \multirow{2}[1]{*}{VSCD} & VSCDNet & \underline{38.1} & \underline{36.9} & \underline{33.9} & \underline{40.7} & \underline{31.7} & \underline{32.1} & \underline{37.7} & \underline{39.0} & \underline{36.6} & \textbf{32.0} & \underline{21.5} & \underline{25.4} \\
        & Ours$^*$ & \textbf{48.07} & \textbf{50.04} & \textbf{46.44} & \textbf{53.29} & \textbf{42.99} & \textbf{42.39} & \textbf{51.10} & \textbf{49.35} & \textbf{48.88} & \underline{26.90} & \textbf{38.13} & \textbf{34.92} \\
        \bottomrule
    \end{tabular}
\end{table}
}

\newcommand{\originalChamgeSimTable}{
\begin{table}[tbh]
    \centering
    \scriptsize
    \caption{\textbf{Performance on ChangeSim under original protocol.}}
    \label{tab:original_changesim}
    \begin{tabular}{lcccc}
        \toprule
        \multirow{2}[1]{*}{\textbf{Method}} & \multicolumn{4}{c}{ChangeSim} \\
        \cmidrule(lr){2-5}
        & Static & Change & mIoU & F1 \\
        \midrule
        FC-EF        & 65.68 & 16.58 & 41.13 & 27.21 \\
        FC-Siam-diff & 84.56 & 25.03 & 54.80 & 38.55 \\
        FC-Siam-conc & 80.63 & 23.32 & 51.95 & 35.44 \\
        CSCDNet      & 90.61 & 24.05 & 57.33 & 35.70 \\
        DR-TANet     & 89.14 & 28.02 & 58.59 & 39.71 \\
        C-3PO        & 91.16 & 28.67 & 59.99 & 40.17 \\
        ZSSCD$^\dagger$        & \underline{92.55} & 28.25 & 60.40 & 39.54 \\
        RobustSCD    & 87.12 & 28.85 & 57.82 & 41.41 \\
        GeSCF$^\dagger$        & 88.82 & 20.27 & 54.54 & 33.28 \\
        TERDNet      & 91.37 & \underline{34.20} & \underline{62.78} & \underline{46.01} \\
        \midrule
        \textbf{Ours} & \textbf{93.87} & \textbf{51.75} & \textbf{72.81} & \textbf{65.13} \\
        \bottomrule
    \end{tabular}
\end{table}
}

\newcommand{\layerAblationTable}{
\begin{table}[tbh]
    \centering
    \scriptsize
    \caption{Layer Ablation}
    \label{tab:layer_ablation}
    \begin{tabular}{lcccclcccc}
        \toprule
        \multirow{2}{*}{\textbf{Layer Configuration}} & \multicolumn{4}{c}{ChangeSim} & \multicolumn{4}{c}{VL-CMU-CD} \\
        \cmidrule(lr){2-5} \cmidrule(lr){6-9}
         & S & C & mIoU & F1 & S & C & mIoU & F1 \\
        \midrule
        Full Fusion &  &  & 72.48 & \underline{60.00} & & & \underline{86.28} & \underline{82.39} \\
        No Layer \#4 & &  & \textbf{72.61} & 59.49 & & & 85.79 & 82.02  \\
        No Layer \#11 &  &  & \underline{72.60} & \textbf{60.88} & & & 85.65 & 80.98 \\
        No Layer \#17 &  &  & 71.59 & 58.07 & & & 86.22 & 82.08 \\
        No Layer \#23 &  &  & 72.46 & 59.68 & & & \textbf{86.40} & \textbf{82.81} \\
        \bottomrule
    \end{tabular}
\end{table}
}

\newcommand{\runtimeTable}{
\begin{table}[!thb]
    \centering
    \scriptsize
    \caption{\textbf{Inference runtime vs.\ number of input images.}}
    \label{tab:runtime}
    \begin{tabular}{lccccccc}
        \toprule
        \multirow{2}{*}{\textbf{Method}} & \multicolumn{7}{c}{\textbf{Number of input images $N$}} \\
        \cmidrule(lr){2-8}
         & 2 & 4 & 8 & 16 & 32 & 64 & 128 \\
        \midrule
        TERDNet & 0.076 & 0.095 & 0.13 & 0.26 & 0.51 & 1.03 & 2.05 \\
        C-3PO & 0.008 & 0.010 & 0.015 & 0.026 & 0.049 & 0.10 & 0.21 \\
        VSCDNet & 0.030 & 0.059 & 0.12 & 0.25 & 0.48 & 0.96 & 1.91 \\
        SceneDiff$^\dagger$ & 22.8 & 32.6 & 51.6 & 96.6 & 210 & 485 & OOM \\
        O-SCD$^\ddagger$ & Fail & Fail & Fail & 187 & 209 & 194 & 286 \\
        \midrule
        Ours-foundation & 0.037 & 0.043 & 0.068 & 0.12 & 0.25 & 0.59 & 1.45 \\
        \bottomrule
    \end{tabular}
\end{table}
}

\newcommand{\rel}[1]{\,($#1$\%)}

\newcommand{\illuminationTable}{
\begin{table}[tbh]
    \centering
    \tiny
    \setlength{\tabcolsep}{1.8pt}
    \caption{\textbf{Simulated illumination changes.} \modelName{} is robust to mild lighting changes and degrades most under extreme brightening. Numbers in parentheses are relative changes (\%) with respect to the unaltered input (None).}
    \label{tab:illumination}
    \begin{tabular}{lcccccccc}
        \toprule
        \multirow{2}[1]{*}{\textbf{Query Illumination}} & \multicolumn{4}{c}{VSCD**} & \multicolumn{4}{c}{\modelName{}-CD-real} \\
        \cmidrule(lr){2-5}\cmidrule(lr){6-9}
         & Static & Change & mIoU & F1 & Static & Change & mIoU & F1 \\
        \midrule
        None           & 98.05 & 68.51 & 83.28 & 53.33 & 97.08 & 47.74 & 72.41 & 41.24 \\
        Mild Dark      & 98.03\rel{0.0} & 68.30\rel{-0.3} & 83.16\rel{-0.1} & 53.20\rel{-0.2} & 97.09\rel{0.0} & 47.90\rel{+0.3} & 72.50\rel{+0.1} & 41.13\rel{-0.3} \\
        Extreme Dark   & 97.88\rel{-0.2} & 66.66\rel{-2.7} & 82.27\rel{-1.2} & 52.45\rel{-1.7} & 97.10\rel{0.0} & 46.96\rel{-1.6} & 72.03\rel{-0.5} & 40.10\rel{-2.8} \\
        Mild Bright    & 97.96\rel{-0.1} & 67.49\rel{-1.5} & 82.72\rel{-0.7} & 53.01\rel{-0.6} & 97.05\rel{0.0} & 47.19\rel{-1.2} & 72.12\rel{-0.4} & 40.95\rel{-0.7} \\
        Extreme Bright & 97.67\rel{-0.4} & 64.39\rel{-6.0} & 81.03\rel{-2.7} & 51.76\rel{-2.9} & 96.72\rel{-0.4} & 44.06\rel{-7.7} & 70.39\rel{-2.8} & 37.82\rel{-8.3} \\
        \bottomrule
    \end{tabular}
\end{table}
}

\newcommand{\slamCDTable}{
    \begin{subtable}[t]{0.55\linewidth}
        \centering
        \caption{Change detection on keyframes.}
        \label{tab:slam_cd}
        \begin{tabular}{lccccc}
            \toprule
            \textbf{Scope} & Pairs & Static & Change & mIoU & F1 \\
            \midrule
            Room  & 78 & 98.58 & 76.56 & 87.57 & 65.06 \\
            House & 12 & 97.14 & 49.53 & 73.33 & 44.66 \\
            \midrule
            All   & 90 & 98.15 & 68.86 & 83.51 & 59.05 \\
            \bottomrule
        \end{tabular}
    \end{subtable}
}

\newcommand{\slamGhostTable}{
    \begin{subtable}[t]{0.42\linewidth}
        \centering
        \caption{Ghost points per state pair.}
        \label{tab:slam_ghost}
        \begin{tabular}{lccc}
            \toprule
            \textbf{Scope} & Naive & \modelName{} & Removed \\
            \midrule
            Room  & 42.1K & 22.9K & 41.9\% \\
            House & 53.7K & 45.1K & 14.1\% \\
            \midrule
            All   & 43.6K & 25.9K & 38.2\% \\
            \bottomrule
        \end{tabular}
    \end{subtable}
}

\newcommand{\slamQuantTable}{
\begin{table}[tbh]
    \centering
    \scriptsize
    \caption{\textbf{\modelName{}-SLAM system-level results on the \modelName{}-CD-SceneSmith test set.} (a) Change detection with the full system agrees with the benchmark evaluation. (b) \modelName{}-SLAM removes far more changed geometry (ghost points) from the map than naively fusing both visits.}
    \label{tab:slam_quant}
    \slamCDTable\hfill\slamGhostTable
\end{table}
}

\newcommand{\slamRealWorldSkipTable}{
\begin{table}[tbh]
    \centering
    \scriptsize
    \caption{\textbf{Skipped keyframes on the real-world sequences.} Skipped keyframes are rare in real-world deployment, even on the largest sequence processed at real-time rate.}
    \label{tab:slam_realworld_skip}
    \begin{tabular}{llccccc}
        \toprule
        \multirow{2}[1]{*}{\textbf{Sequence}} & \multirow{2}[1]{*}{\textbf{Capture}} & \multicolumn{2}{c}{Prior visit} & \multicolumn{2}{c}{Current visit} & Both visits \\
        \cmidrule(lr){3-4}\cmidrule(lr){5-6}\cmidrule(lr){7-7}
         & & Keyframes & Skipped & Keyframes & Skipped & Skipped \\
        \midrule
        Office          & Handheld & 55  & 1.8\% & 21  & 0\%   & 1.3\% \\
        Home            & Handheld & 103 & 0\%   & 83  & 0\%   & 0\%   \\
        Atrium          & Robot    & 283 & 0\%   & 324 & 0.6\% & 0.3\% \\
        Workshop        & Robot    & 180 & 0\%   & 159 & 0\%   & 0\%   \\
        Full-floor loop & Robot (real-time) & 841 & 4.0\% & 805 & 8.4\% & 6.2\% \\
        \bottomrule
    \end{tabular}
\end{table}
}
\newcommand{\teaserFigure}{
\begin{figure}[h]
    \centering
    \includegraphics[width=0.9\linewidth]{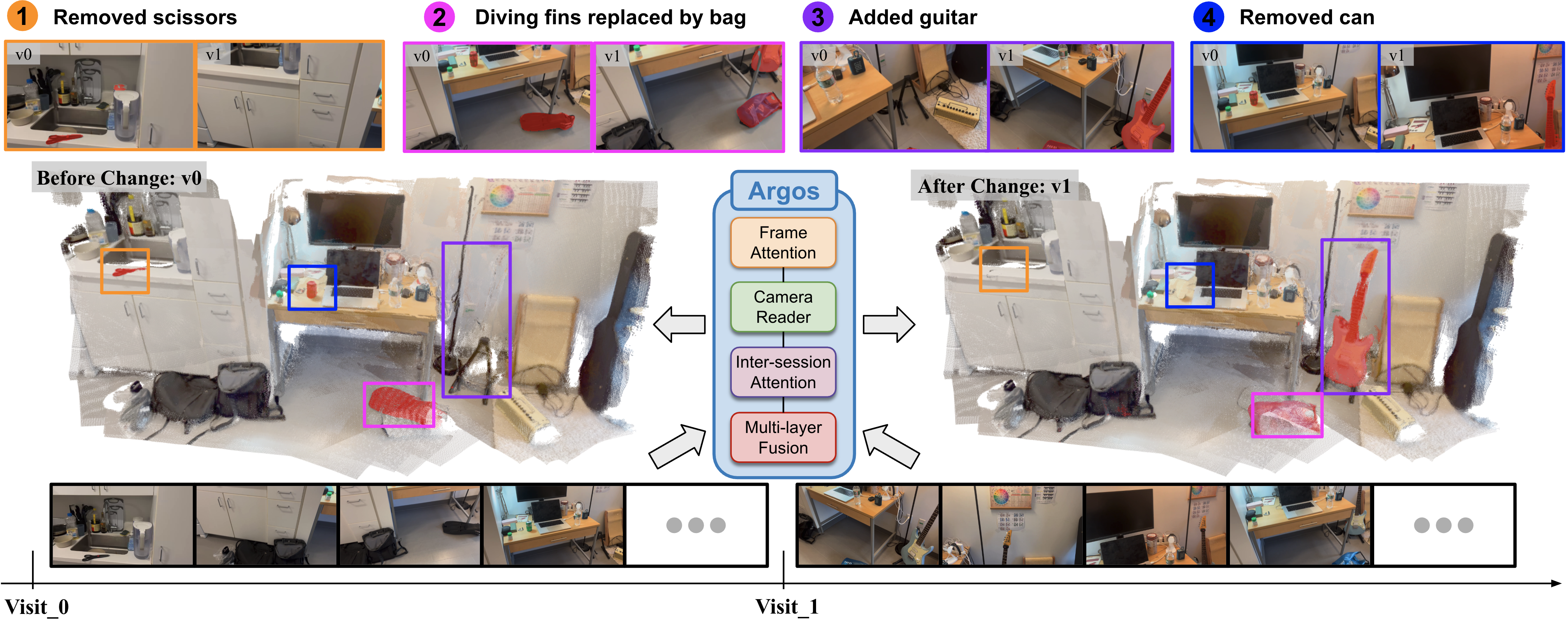}
    \caption{We propose \textbf{\modelName{}}, a feedforward model for sub-second change detection and 3D reconstruction, and integrate it into \modelName{}-SLAM for online 4D reconstruction of change-aware point-cloud maps. The figure illustrates how \modelName{} captures scene changes, including the added guitar (3) and removed can (4), while maintaining a consistent spatio-temporal representation across visits.}
    \label{fig:teaser}
\end{figure}
}

\newcommand{\LayerAttentionViz}{
\begin{figure}[t]
    \centering
    \includegraphics[width=0.9\linewidth]{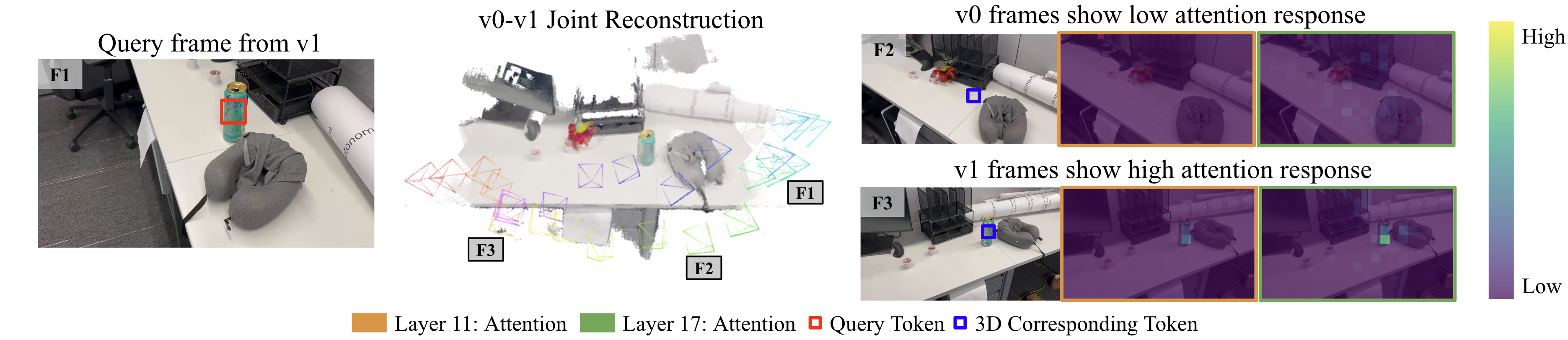}
    \caption{\textbf{Attention score visualization of selected layers.} The token associated with the added green bottle in $v_1$ shows strong attention to the corresponding bottle region across different views within the same visit, while finding no high-attention score in $v_0$, where the bottle is absent.}
    \label{fig:layer_attention_viz}
\end{figure}
}

\newcommand{\architectureFigure}{
\begin{figure}[t]
    \centering
    \includegraphics[width=0.9\linewidth]{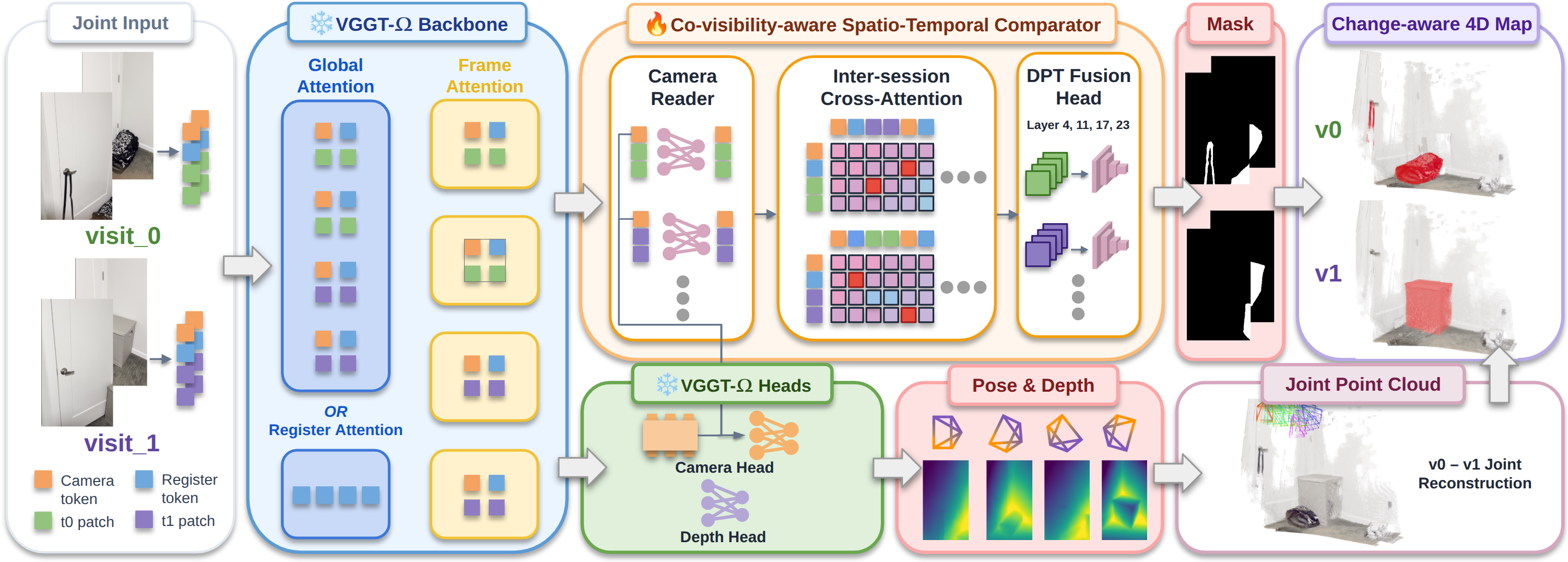}
    \caption{\textbf{\modelName{} Architecture.} Building on VGGT-$\Omega$, we introduce a Co-visibility-aware Spatio-Temporal Comparator with three modules: a Camera Reader that injects final camera tokens into intermediate patch features, Inter-session Cross-Attention that aggregates information across visits, and a lightweight DPT Fusion Head that combines multi-layer features to predict per-visit change masks. These masks are then used to disentangle the joint reconstruction into a change-aware map.}
    \label{fig:architecture}
\end{figure}
}

\newcommand{\QualitativeFigure}{
\begin{figure}[t]
    \centering
    \begin{subfigure}{\textwidth}
        \centering
        \includegraphics[height=0.2cm]{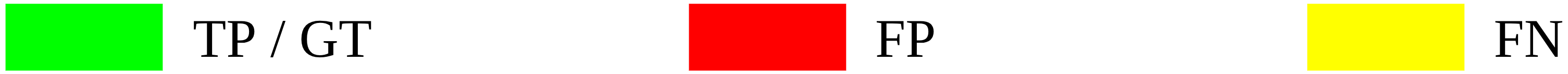}
    \end{subfigure}
    
    \vspace{0.2em}
    
    \begin{subfigure}[b]{0.34\textwidth}
        \centering
        \includegraphics[height=5.6cm]{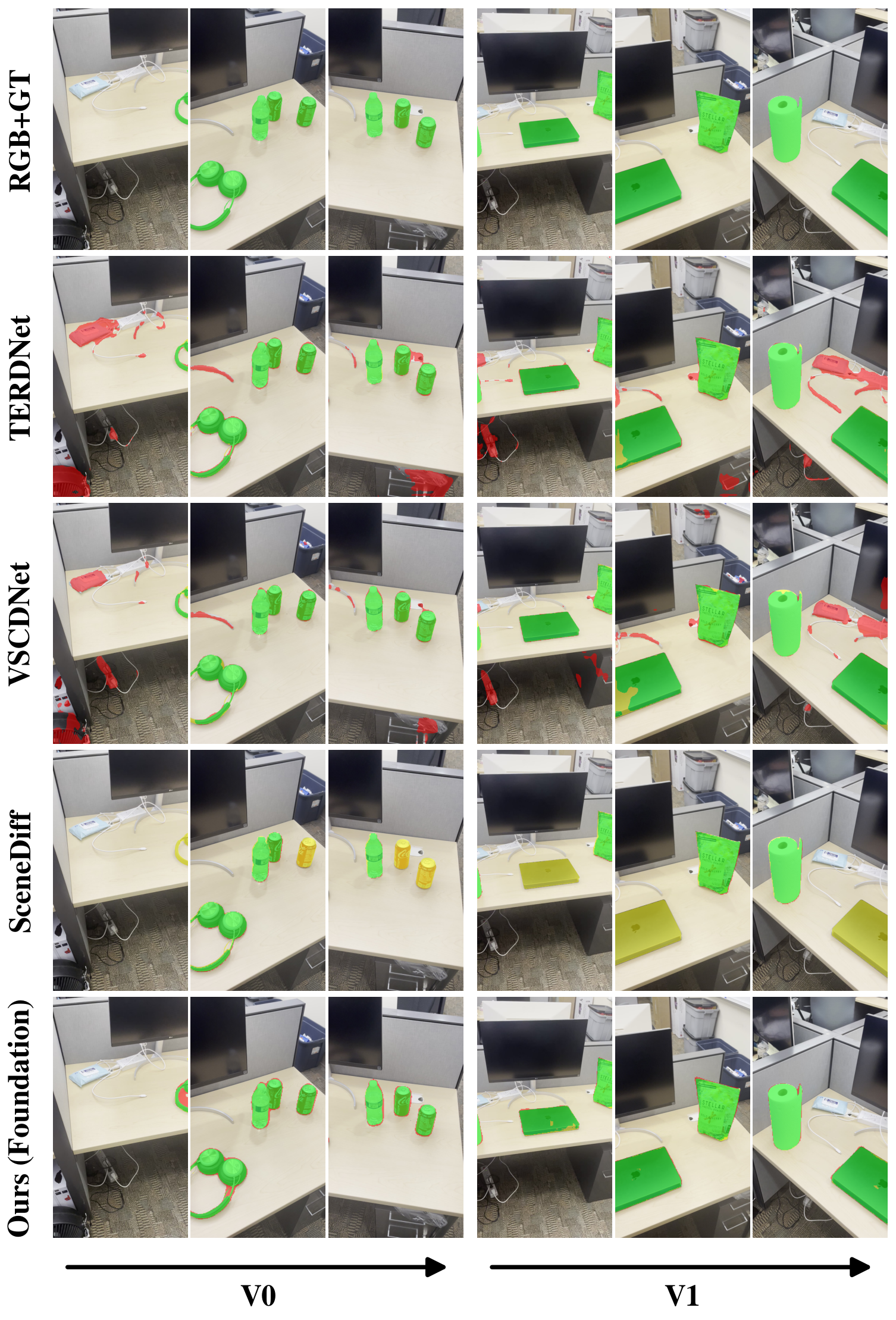}
    \end{subfigure}
    \hfill
    \begin{subfigure}[b]{0.65\textwidth}
        \centering
        \includegraphics[height=5.6cm]{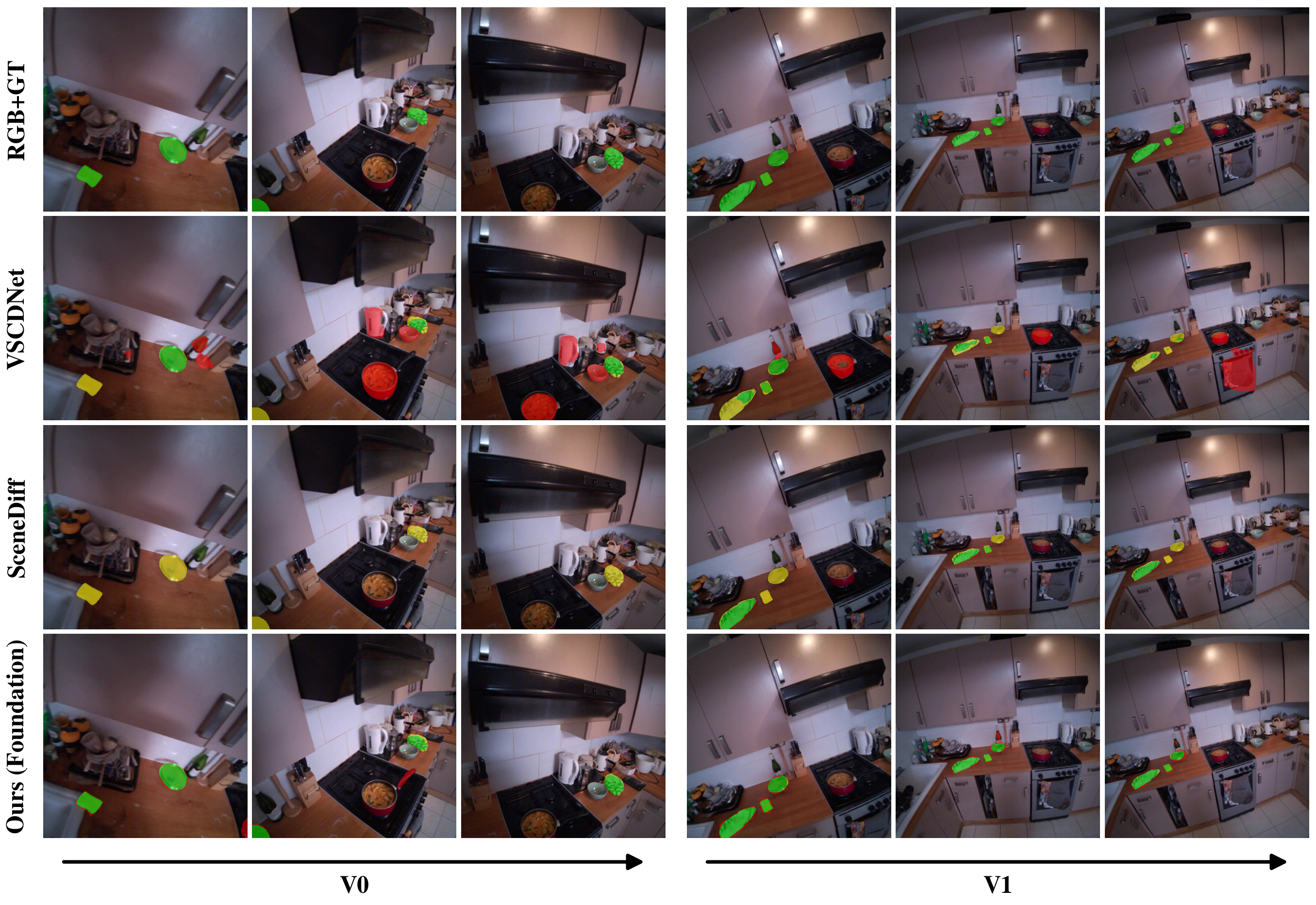}
    \end{subfigure}
    
    \caption{\textbf{Qualitative results.} We show results of the best-performing models. \modelName{} shows cleaner masks and fewer false positives or false negatives in real-world. Appendix~\ref{app:additional_qualitative} shows more results.}
    \label{fig:main_dataset_result}
\end{figure}
}

\newcommand{\realworldFigure}{
\begin{figure}[t]
    \centering
    \includegraphics[width=0.9\linewidth]{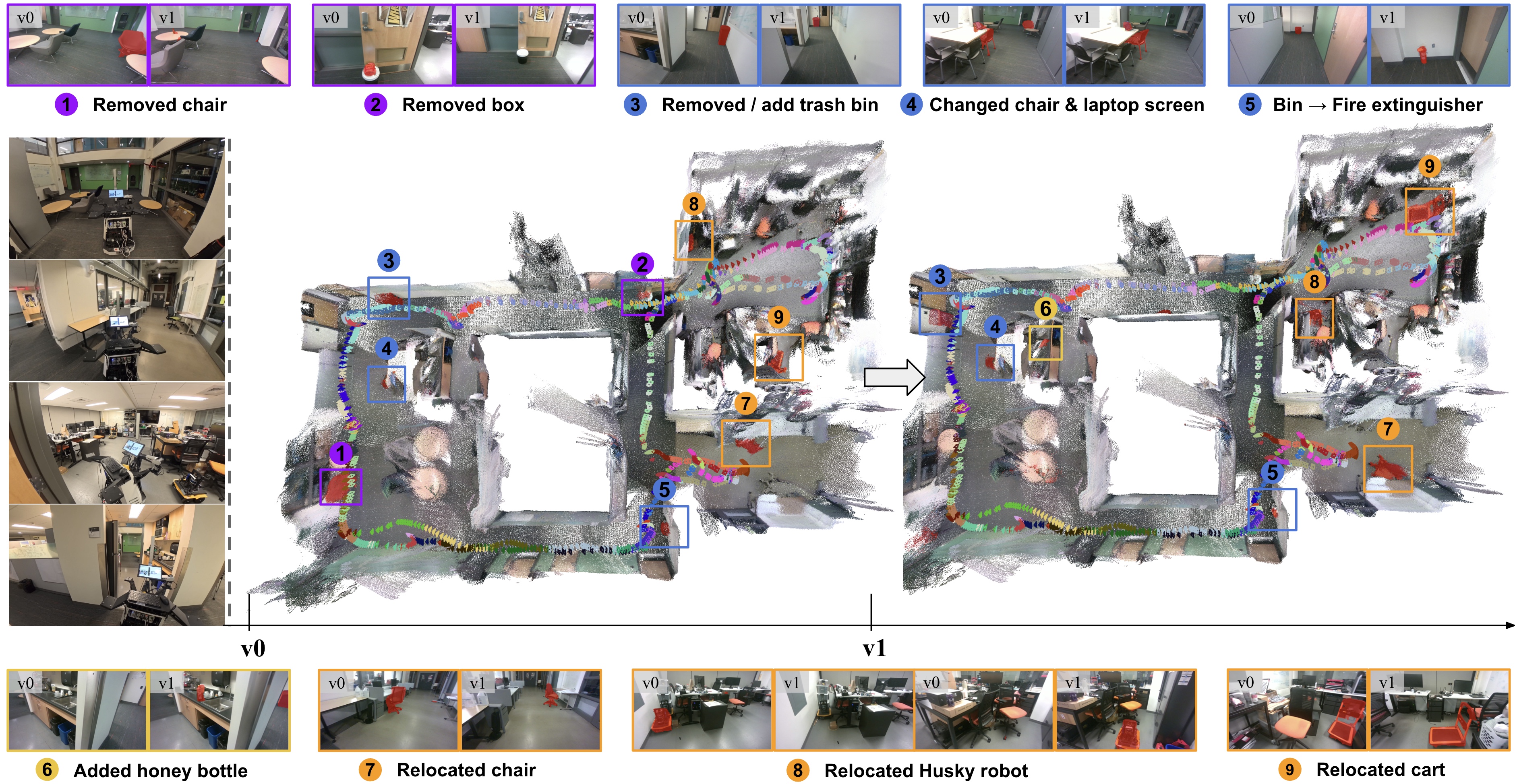}
    \caption{\textbf{Qualitative result of the online system in a multi-room scene.}
\modelName{}-SLAM localizes additions and removals across object scales, such as a removed chair, an added honey bottle, and a relocated Husky robot, consistently in both 2D observations and the global 4D map.}
    \label{fig:real_world_figure}
\end{figure}
}

\newcommand{\failureCaseFigure}{
\begin{figure}[t]
    \centering
    \begin{subfigure}[b]{0.340\textwidth}
        \centering
        \includegraphics[width=\linewidth]{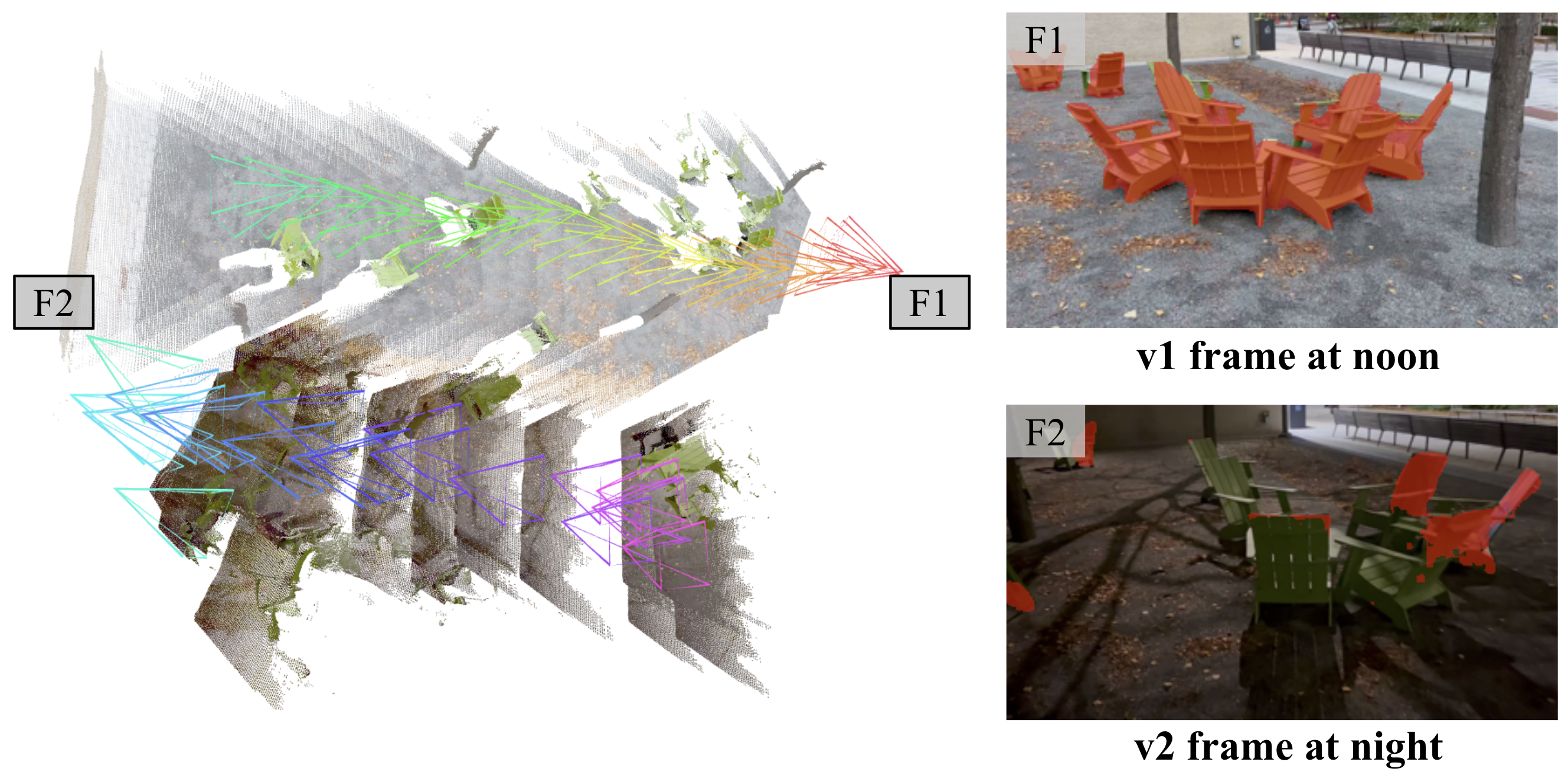}
        \caption{Extreme illumination change.}
        \label{fig:failure_illumination}
    \end{subfigure}
    \hfill
    \begin{subfigure}[b]{0.280\textwidth}
        \centering
        \includegraphics[width=\linewidth]{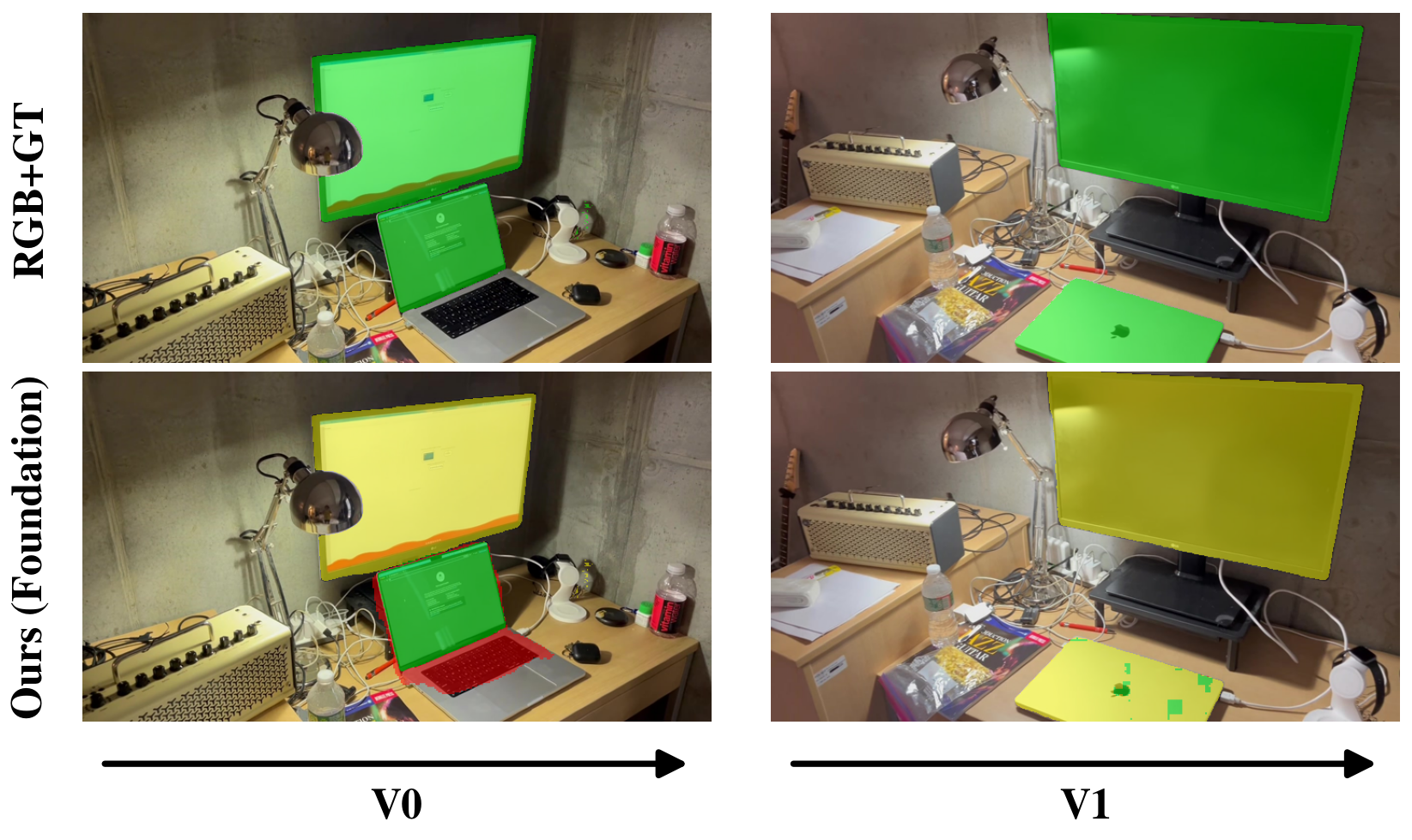}
        \caption{Semantic change.}
        \label{fig:failure_semantic}
    \end{subfigure}
    \hfill
    \begin{subfigure}[b]{0.280\textwidth}
        \centering
        \includegraphics[width=\linewidth]{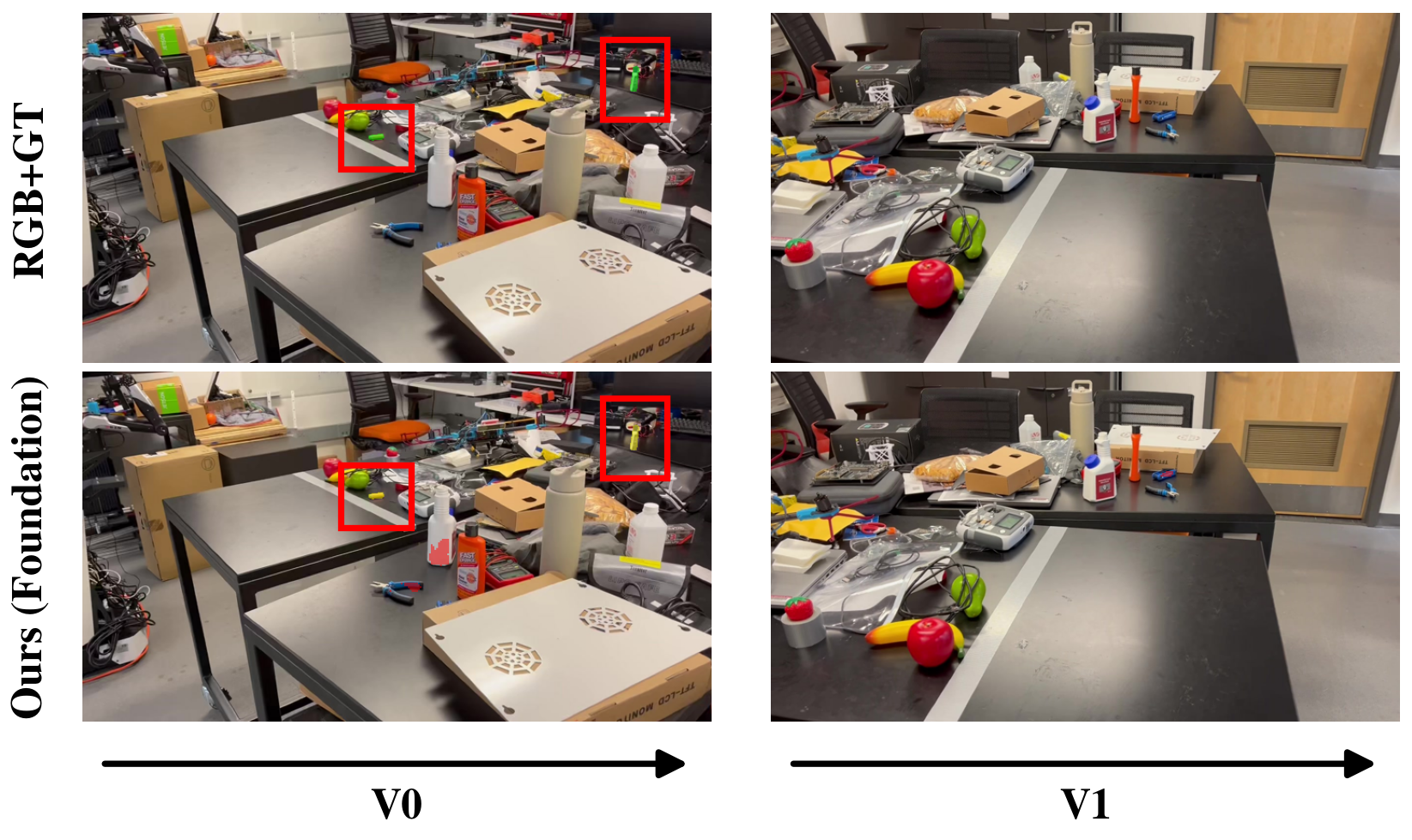}
        \caption{Tiny objects in clutter.}
        \label{fig:failure_tiny}
    \end{subfigure}
    \caption{\textbf{Challenging cases.} \modelName{} fails when visits cannot be registered or changes leave little geometric evidence.}
    \label{fig:failure_cases}
\end{figure}
}

\newcommand{\homeExtremeFailureFigure}{
\begin{figure}[tbh]
    \centering
    \includegraphics[width=\linewidth]{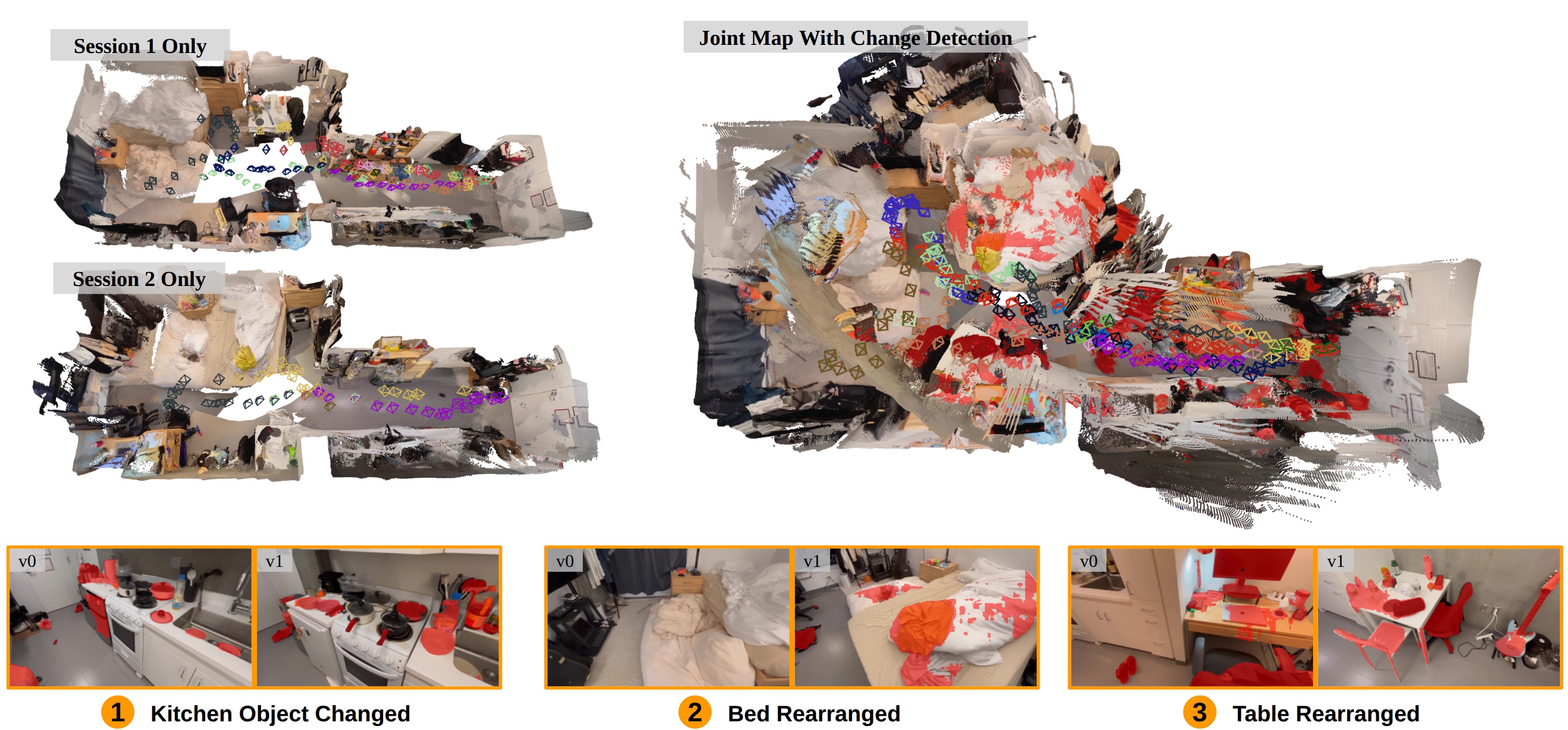}
    \caption{\textbf{\modelName{}-SLAM under heavy furniture rearrangement in a multi-room home.} Top left: the map of each visit reconstructed alone, where Session 1 is $v_0$ and Session 2 is $v_1$. Top right: the joint map, with detected changes in red. Bottom: keyframe pairs from the numbered regions, with predicted changes overlaid.}
    \label{fig:failure_home_extreme}
\end{figure}
}

\newcommand{\datasetGenerationFigure}{
\begin{figure}[htb]
    \centering
    \begin{subfigure}[b]{0.45\textwidth}
        \centering
        \includegraphics[width=1.0\linewidth]{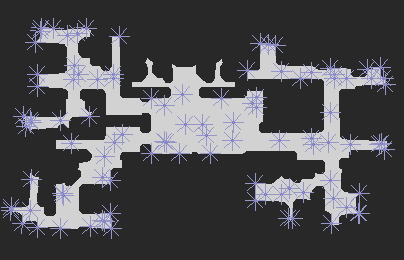}
        \caption{Step 1: initial viewpoint sampling.}
    \end{subfigure}
    \begin{subfigure}[b]{0.45\textwidth}
        \centering
        \includegraphics[width=1.0\linewidth]{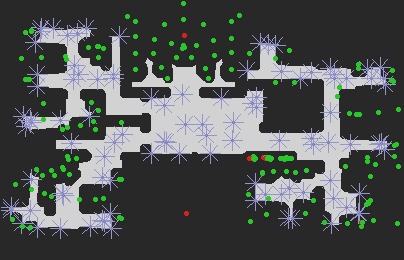}
        \caption{Step 2: viewpoint object coverage check.}
    \end{subfigure}
    
    \vspace{0.2em}
    
    \begin{subfigure}[b]{0.45\textwidth}
        \centering
        \includegraphics[width=1.0\linewidth]{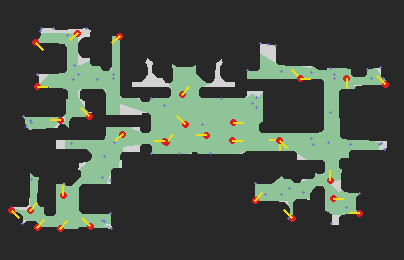}
        \caption{Step 3: fog-of-war viewpoints selection.}
    \end{subfigure}
    \begin{subfigure}{0.45\textwidth}
        \centering
        \includegraphics[width=1.0\linewidth]{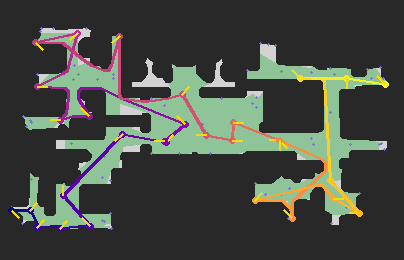}
        \caption{Step 4: TSP routing and trajectory generation.}
    \end{subfigure}
    
    \caption{\textbf{Synthetic dataset trajectory generation process.} The steps ensure maximum coverage of visible objects with a small number of viewpoints to ensure sufficient and efficient coverage of the environments.}
    \label{fig:dataset_generation}
\end{figure}
}

\newcommand{\datasetCovisibilityFigure}{
\begin{wrapfigure}{r}{0.5\textwidth}
    \centering
    \includegraphics[width=1\linewidth]{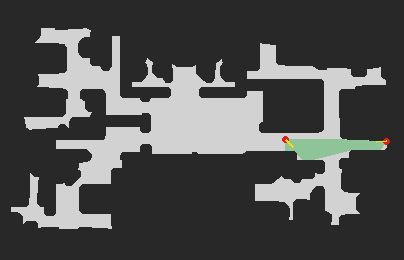}
    \caption{\textbf{Frustum overlap view match.}}
    \label{fig:view_match}
\end{wrapfigure}
}

\newcommand{\oscdComparison}{
\begin{figure}[thb]
    \centering

    \includegraphics[width=1.0\linewidth]{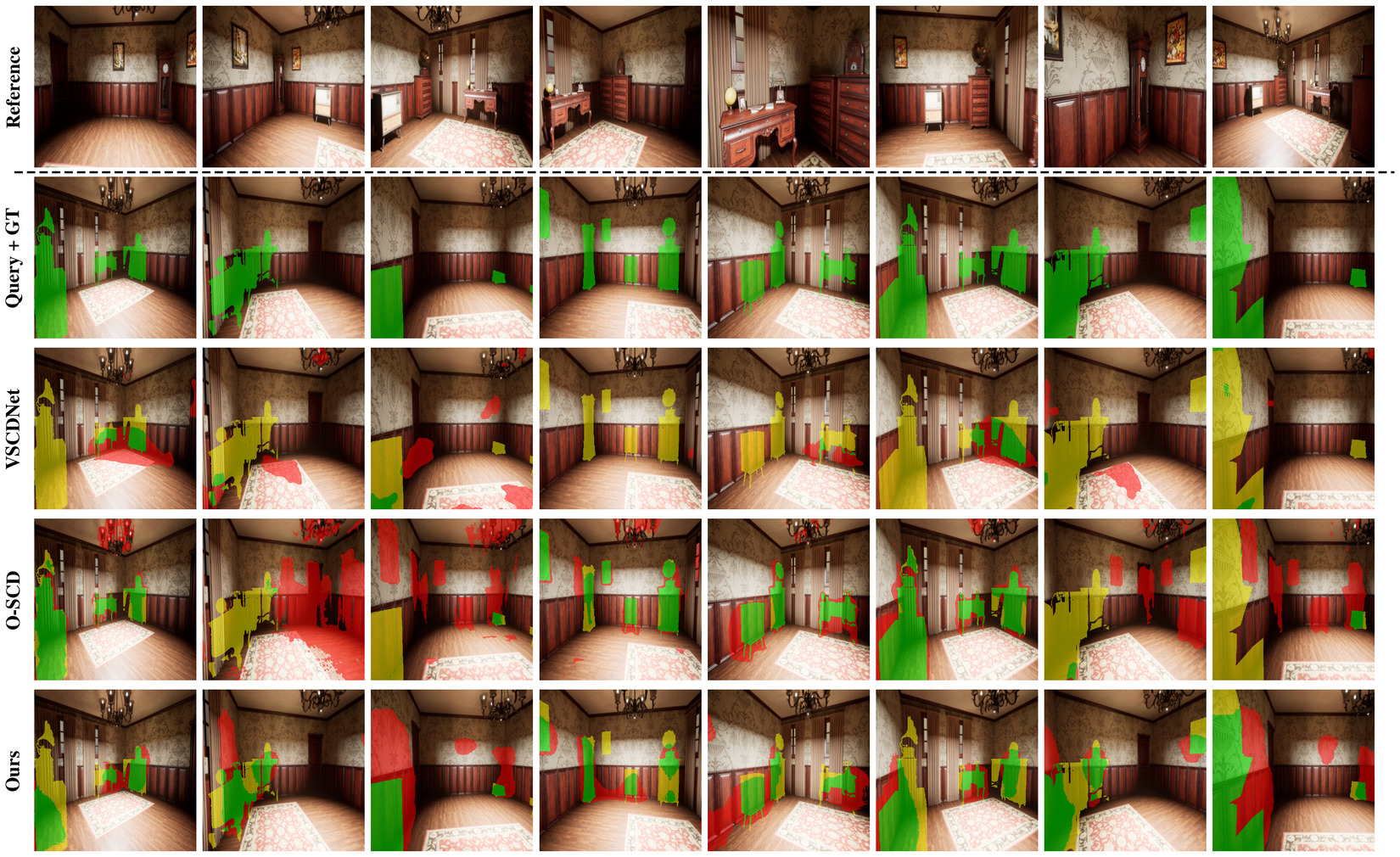}
    \caption{\textbf{Qualitative comparison between O-SCD, VSCDNet, and \modelName{}.} Even with 128 reference frames, O-SCD produces sensible predictions---particularly over removed objects---but still yields many false positives (FP) and false negatives (FN).}
    \label{fig:oscd_qualitative}
\end{figure}
}

\newcommand{\oscdFailureVis}{
\begin{figure}[tbh]
    \centering
    \begin{subfigure}{\textwidth}
        \centering
        \includegraphics[width=1.0\linewidth]{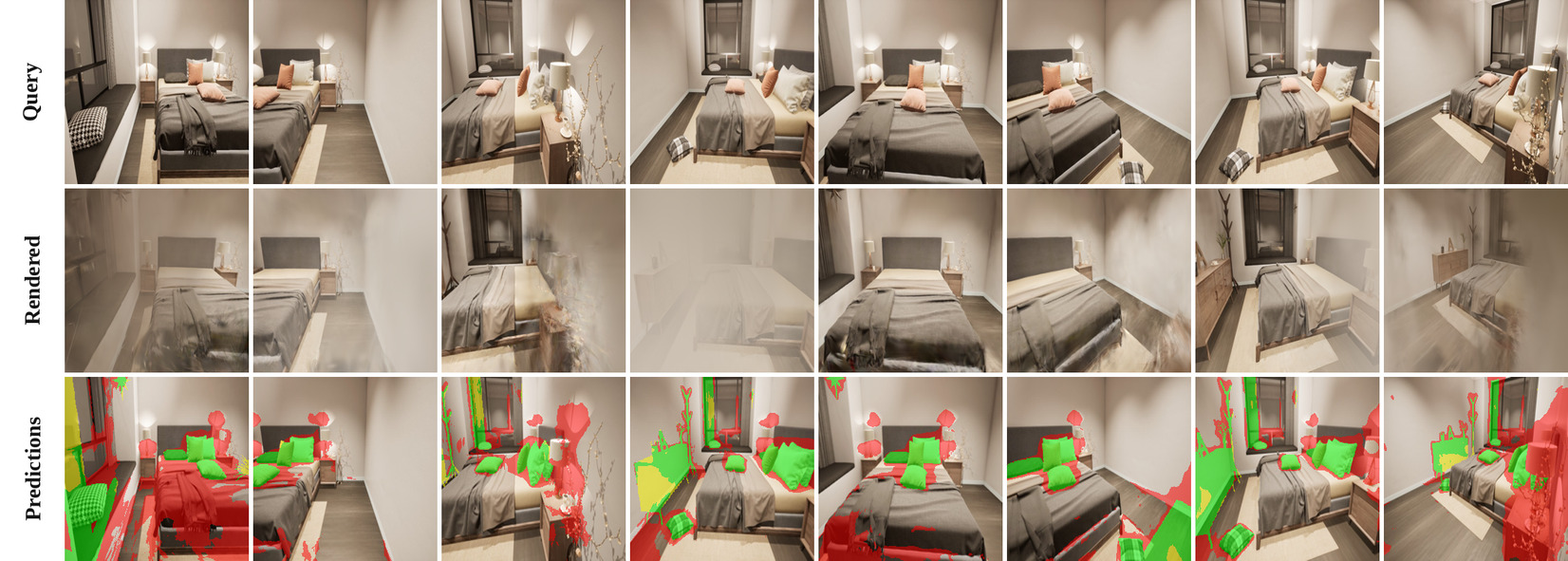}
        \caption{Poor rendering quality from the Gaussian Splats at novel views corresponding to the query camera positions.}
    \end{subfigure}
    \vspace{2px}
    \begin{subfigure}{\textwidth}
        \centering
        \includegraphics[width=1.0\linewidth]{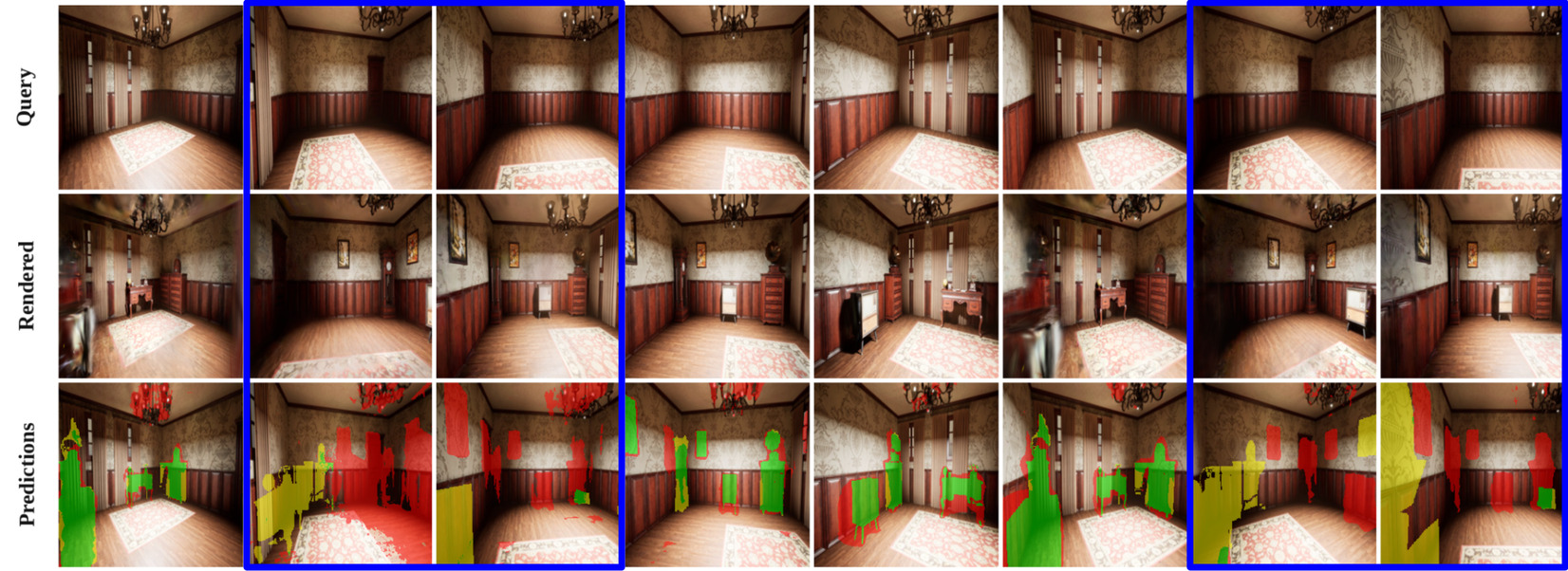}
        \caption{\textcolor{blue}{\textbf{Blue boxes}} highlight incorrect pose estimation, which is readily identifiable from the carpet orientation.}
    \end{subfigure}
    \caption{\textbf{O-SCD limitation analysis.} Even when COLMAP and Gaussian Splatting succeed, poor rendering quality or incorrect camera registration can still lead to noisy predictions.}
    \label{fig:oscd_failure_vis}
\end{figure}
}

\newcommand{\parameterCount}{
\begin{figure}[htbp]
    \centering
    \includegraphics[width=0.75\linewidth]{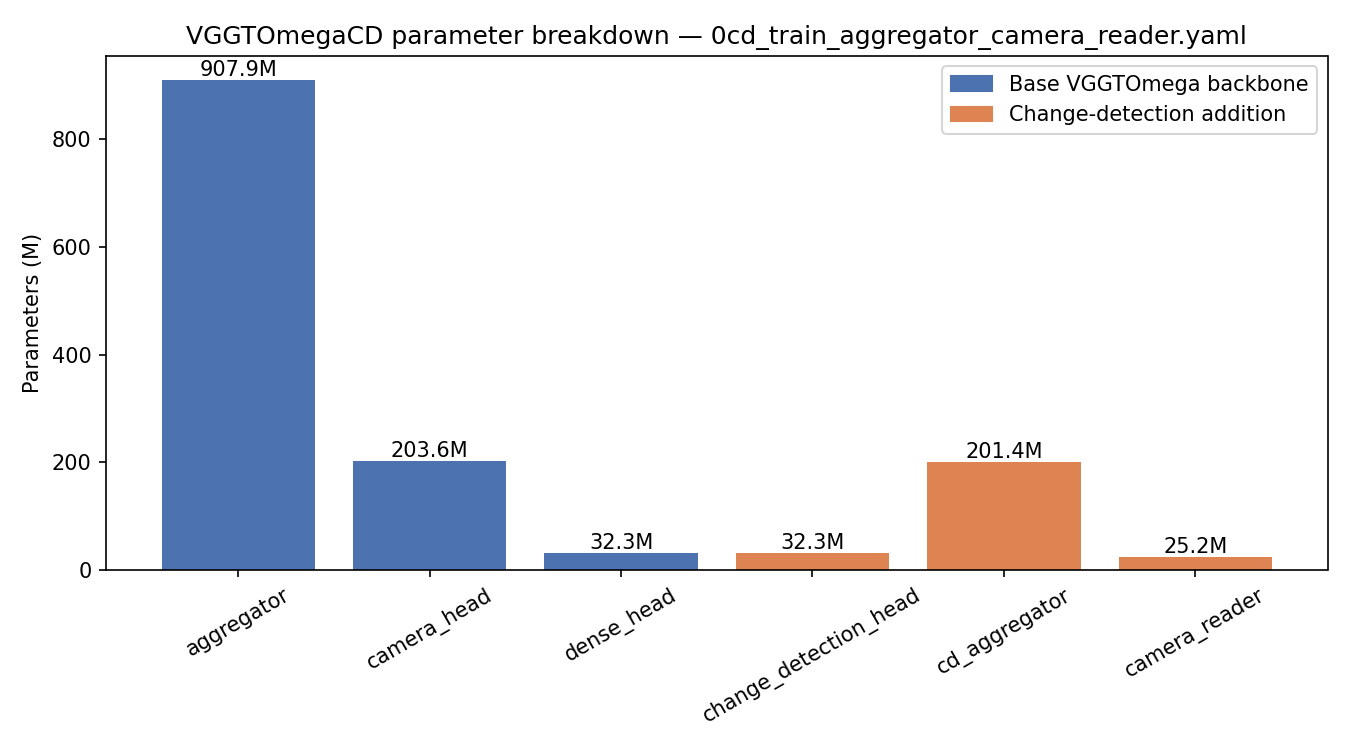}
    \caption{\textbf{Model parameter Count.}}
    \label{fig:parameter_count}
\end{figure}
}

\newcommand{\backboneAttentionFigure}{
\begin{figure}[htb]
    \centering
    \includegraphics[width=\linewidth]{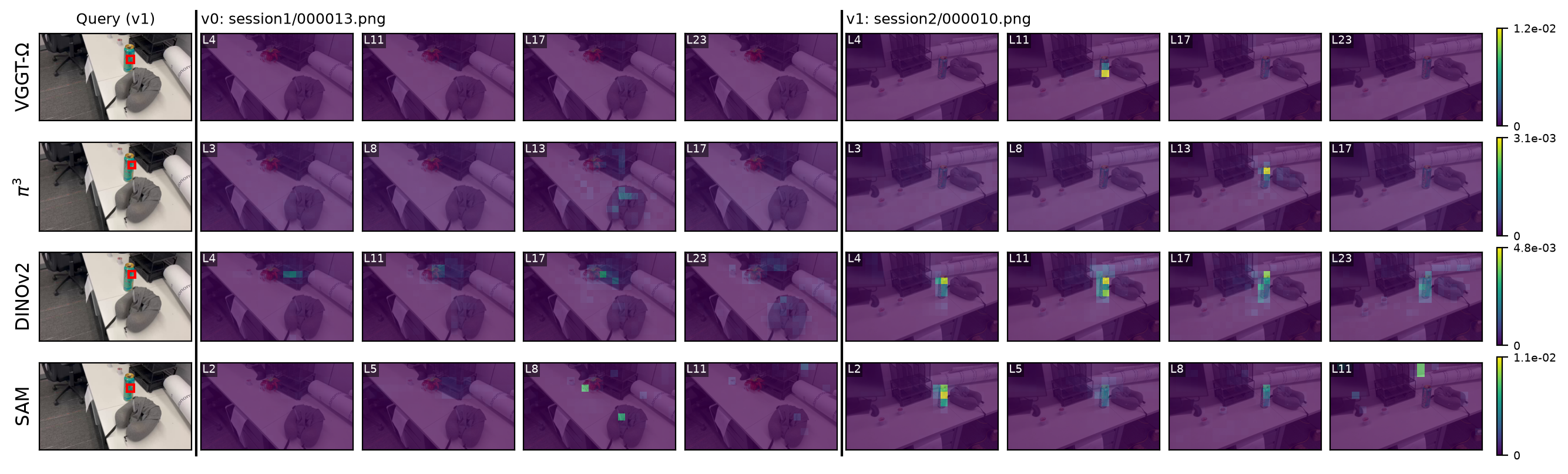}
    \caption{\textbf{Backbone attention comparison.} We query a patch on a can (red box) that appears in $v_1$ but not in $v_0$, and show its attention over one $v_0$ frame and one $v_1$ frame at four layers (L) of each backbone. Each row uses its own color scale, so colors are comparable within a row but not across rows. Ideal attention is high on the can in $v_1$ and near zero everywhere in $v_0$.}
    \label{fig:backbone_attention}
\end{figure}
}

\newcommand{\altriumFigure}{
\begin{figure}[htb]
    \centering
    \includegraphics[width=\linewidth]{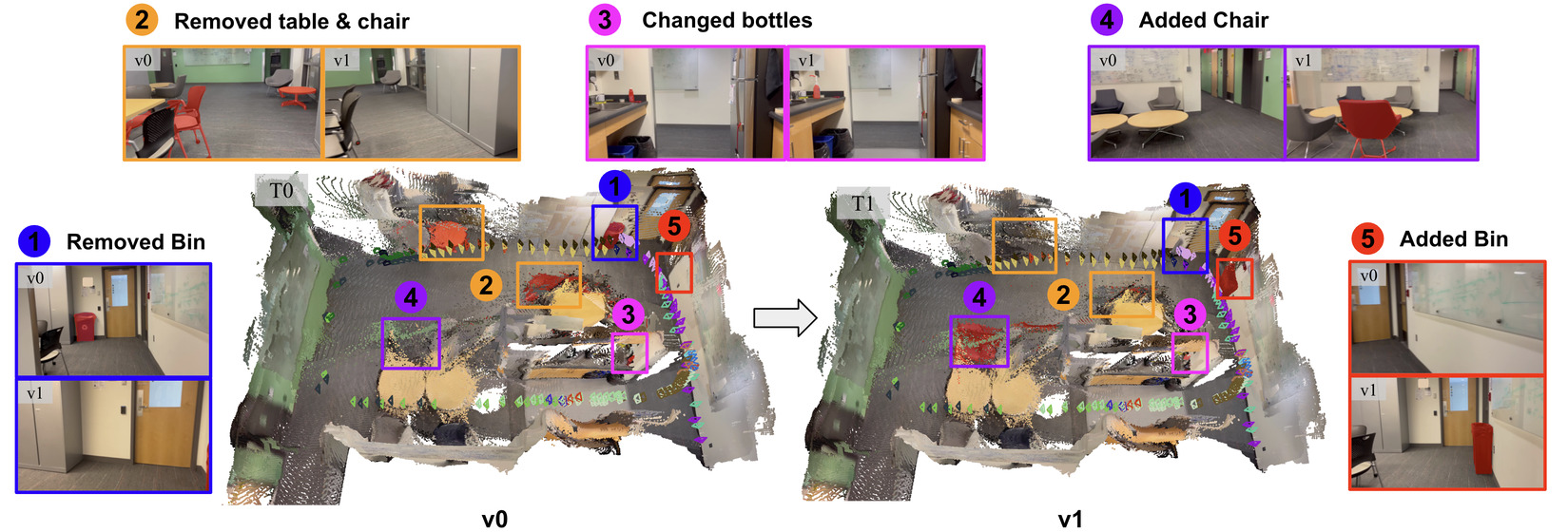}
    \caption{\textbf{Qualitative result of online robot mapping in a medium-sized common space area.}}
    \label{fig:31_altrium}
\end{figure}
}

\newcommand{\robotWorkshopFigure}{
\begin{figure}[htb]
    \centering
    \includegraphics[width=\linewidth]{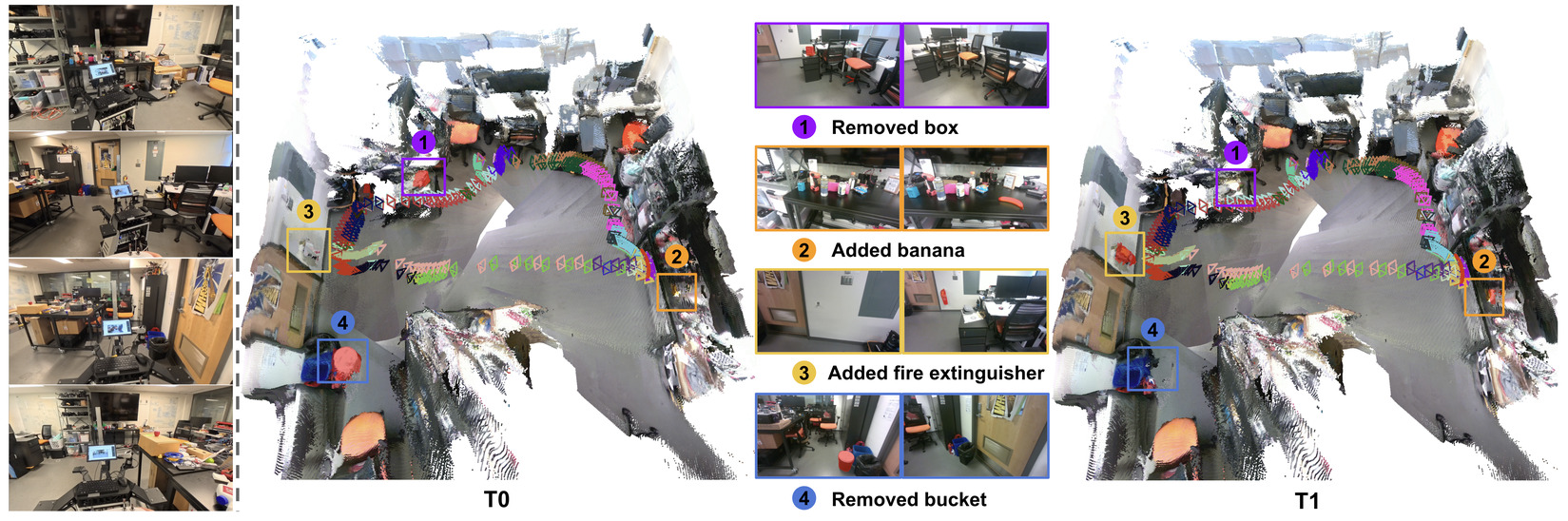}
    \caption{\textbf{Qualitative result of online robot mapping in a small-size workshop environment.}}
    \label{fig:robot_workshop}
\end{figure}
}

\newcommand{\humanHomeFigure}{
\begin{figure}[!htb]
    \centering
    \includegraphics[width=0.9\linewidth]{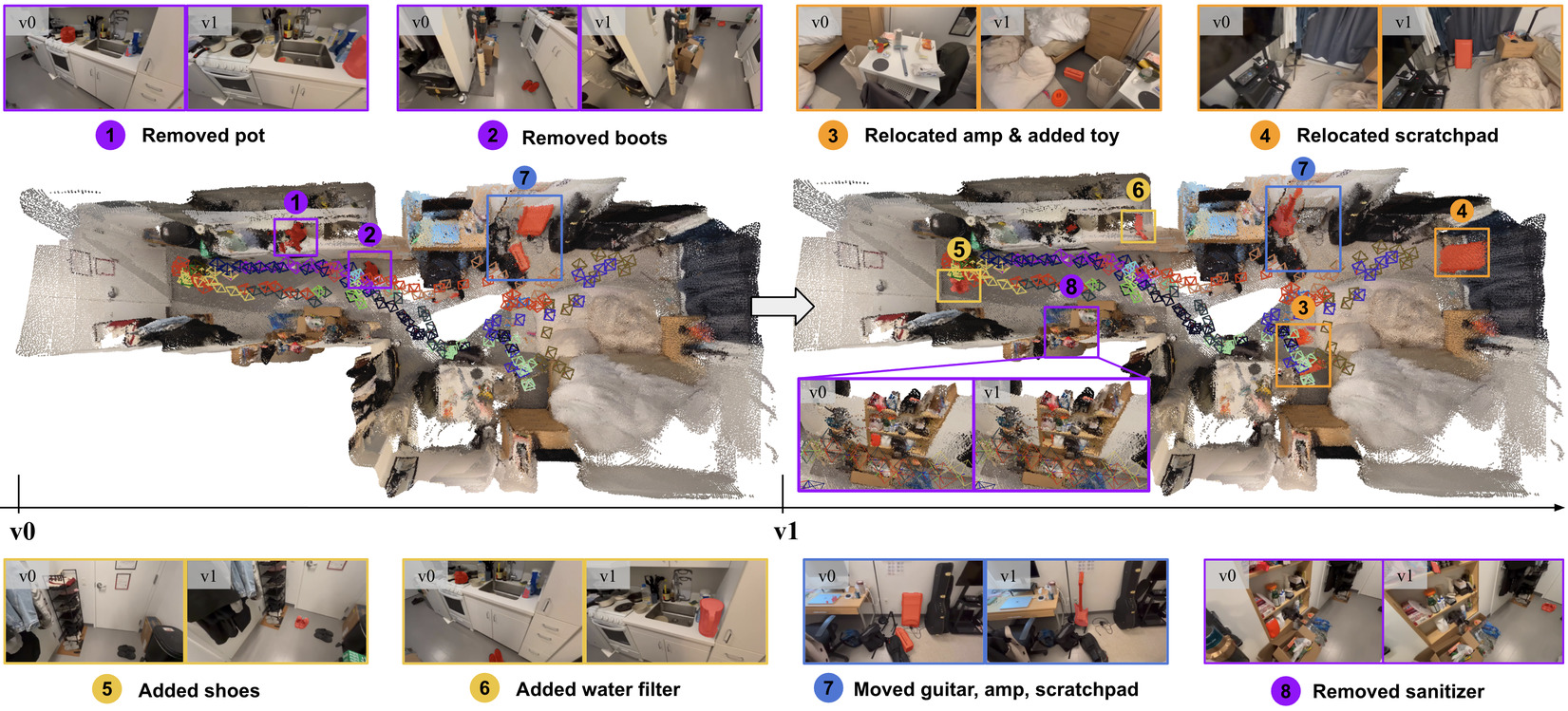}
    \caption{\textbf{Qualitative result of the offline mapping from iPhone-captured video in a multi-room apartment.}}
    \label{fig:human_home}
\end{figure}
}

\newcommand{\humanOfficeFigure}{
\begin{figure}[!htb]
    \centering
    \includegraphics[width=0.9\linewidth]{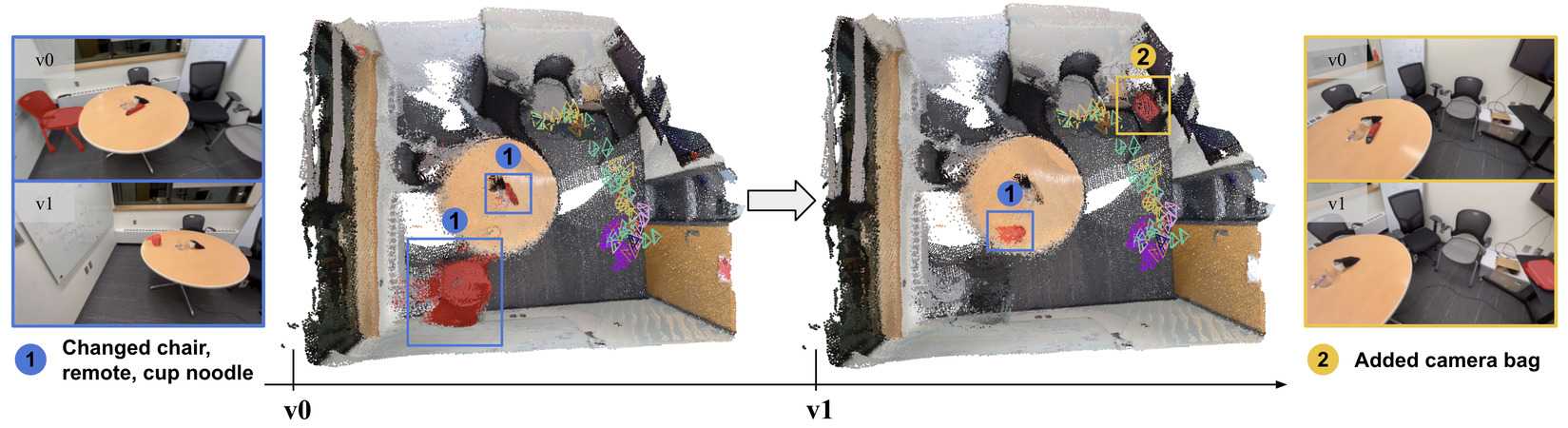}
    \caption{\textbf{Qualitative result of the offline mapping from iPhone-captured video in a small office.}}
    \label{fig:human_office}
\end{figure}
}

\newcommand{\SCDResultFigure}{
\begin{figure}[thb]
    \centering
     \begin{subfigure}[b]{0.29\textwidth}
        \centering
        \includegraphics[width=1.0\linewidth]{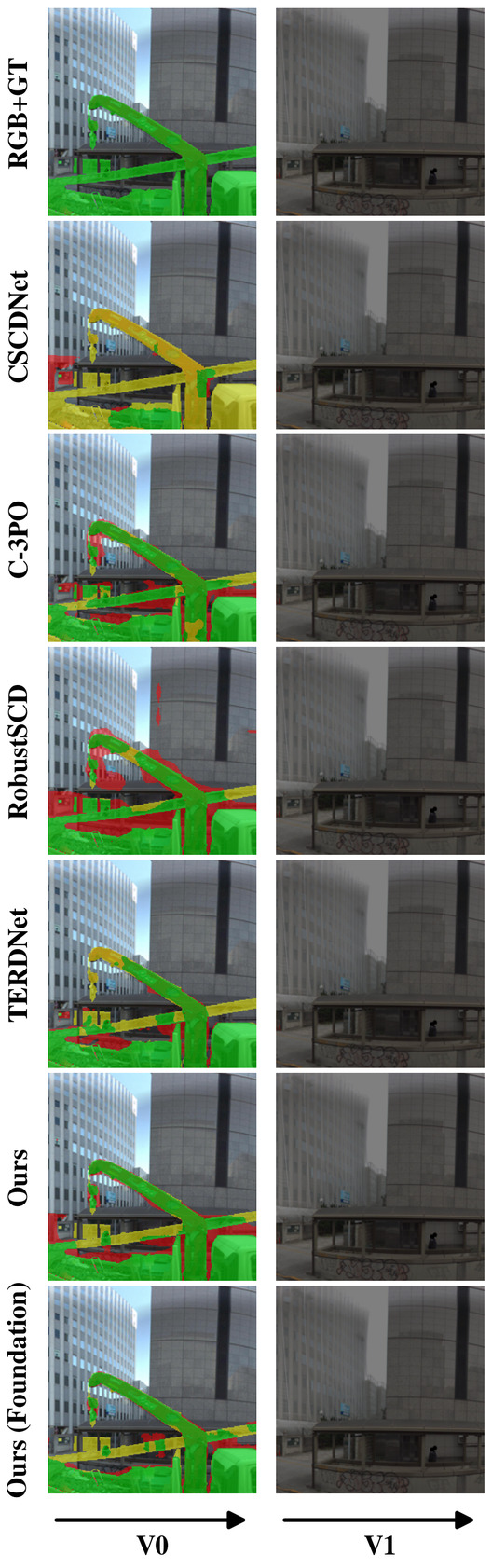}
        \caption{PSCD}
    \end{subfigure}
    \hfill
    \begin{subfigure}[b]{0.29\textwidth}
        \centering
        \includegraphics[width=1.0\linewidth]{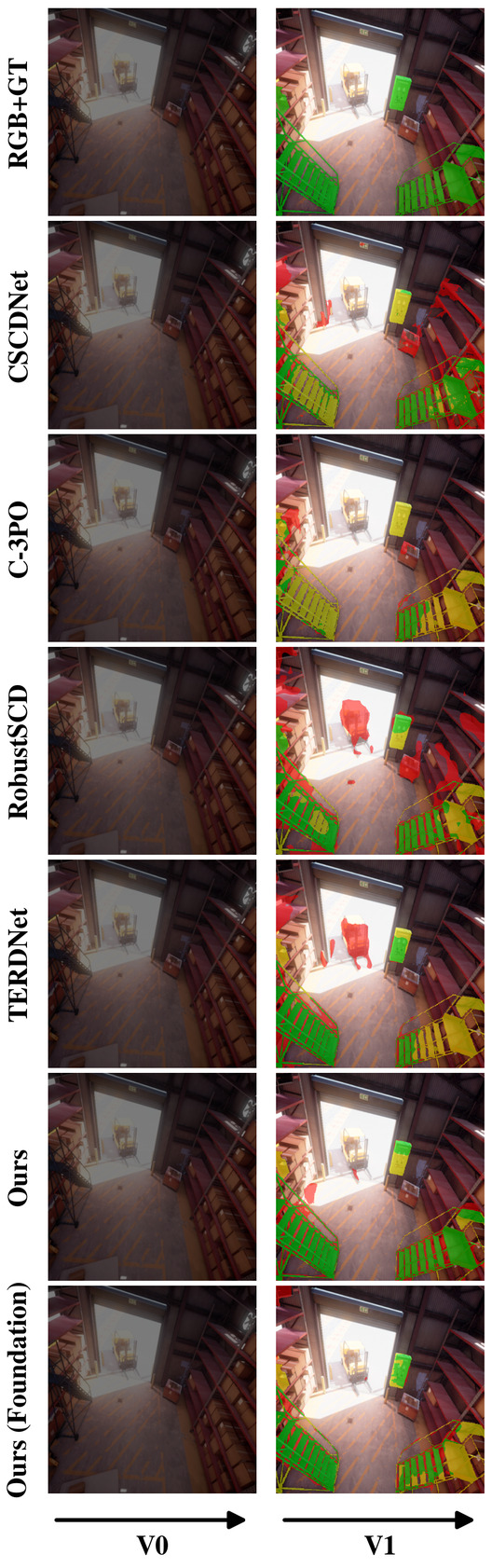}
        \caption{ChangeSim}
    \end{subfigure}
    \hfill
    \begin{subfigure}[b]{0.29\textwidth}
        \centering
        \includegraphics[width=1.0\linewidth]{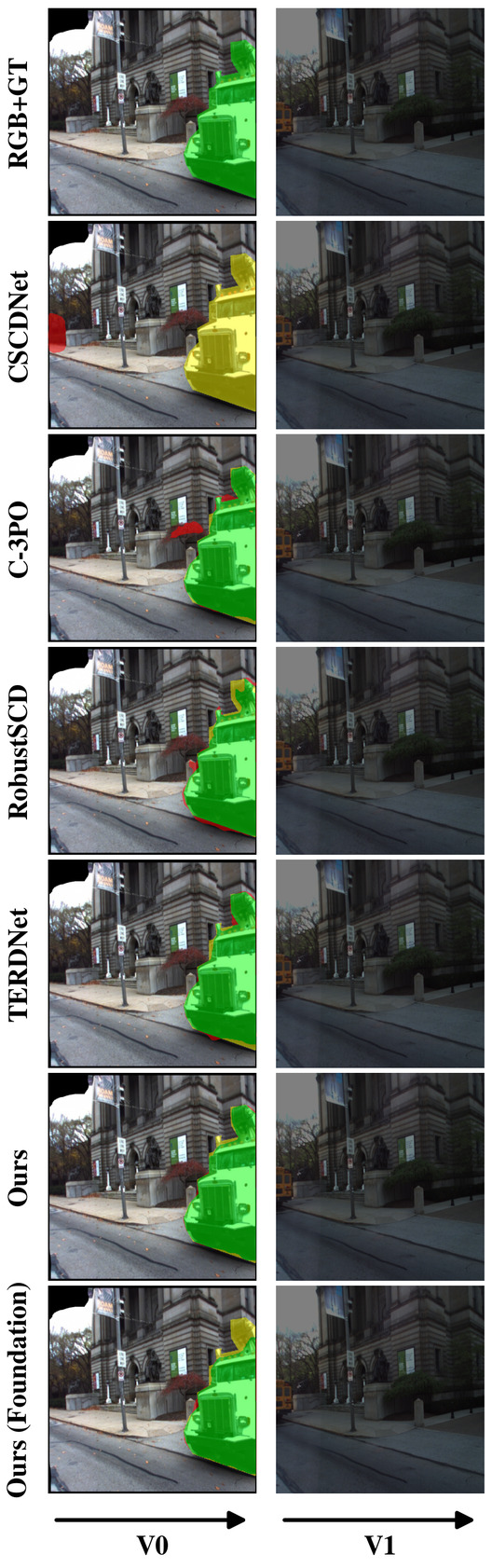}
        \caption{VL-CMU-CD}
    \end{subfigure}
    
    \caption{\textbf{Qualitative results of SCD datasets.} SCD contains image pairs with labels only on the $v_0$ or $v_1$ image. \modelName{} consistently produces predictions with less noise. }
    \label{fig:scd_dataset_result}
\end{figure}
}

\newcommand{\VSCDResultFigure}{

    \begin{figure}[htb]
        \centering
        \includegraphics[width=1\linewidth]{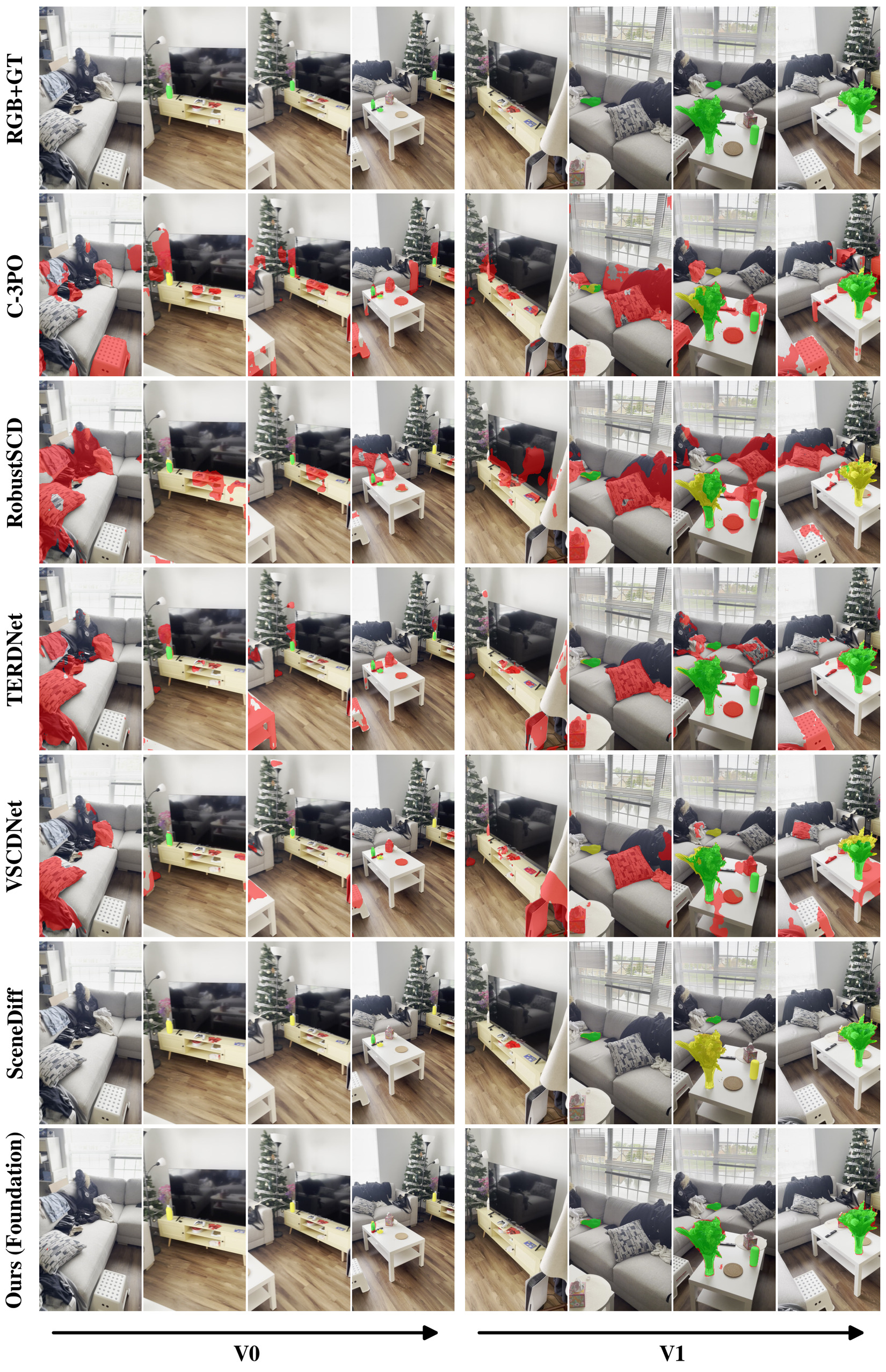}
        \caption{\textbf{Qualitative result of the SceneDiff dataset.}}
        \label{fig:scenediff_result}
    \end{figure}
    
    \begin{figure}[htb]
        \centering
        \includegraphics[width=1\linewidth]{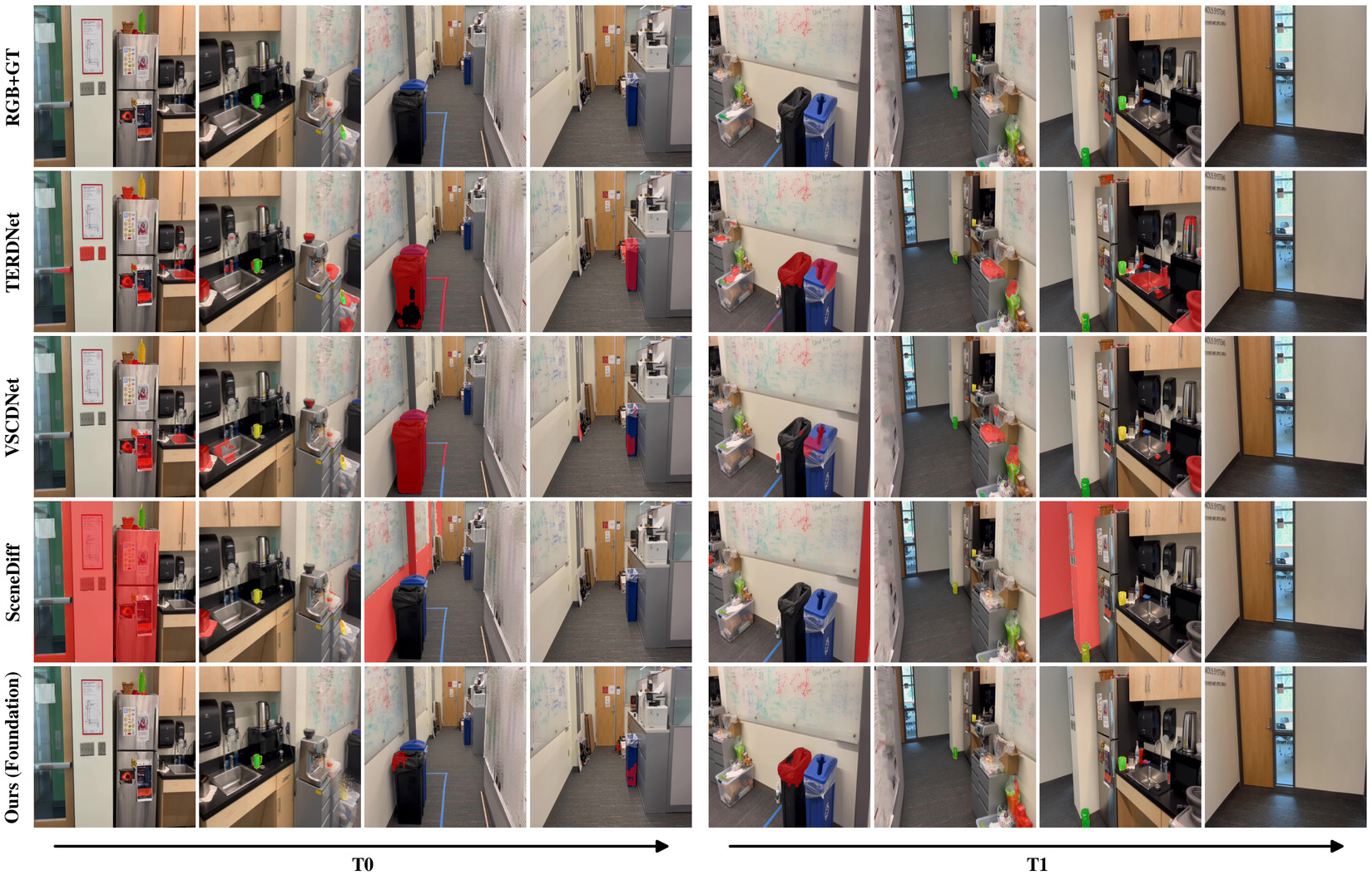}
        \caption{\textbf{Qualitative result of the \modelName{}-CD-real dataset.} Stretched vertically to provide a better view. }
        \label{fig:argos_cd_real_result}
    \end{figure}

    \begin{figure}[htb]
        \centering
        \includegraphics[width=1\linewidth]{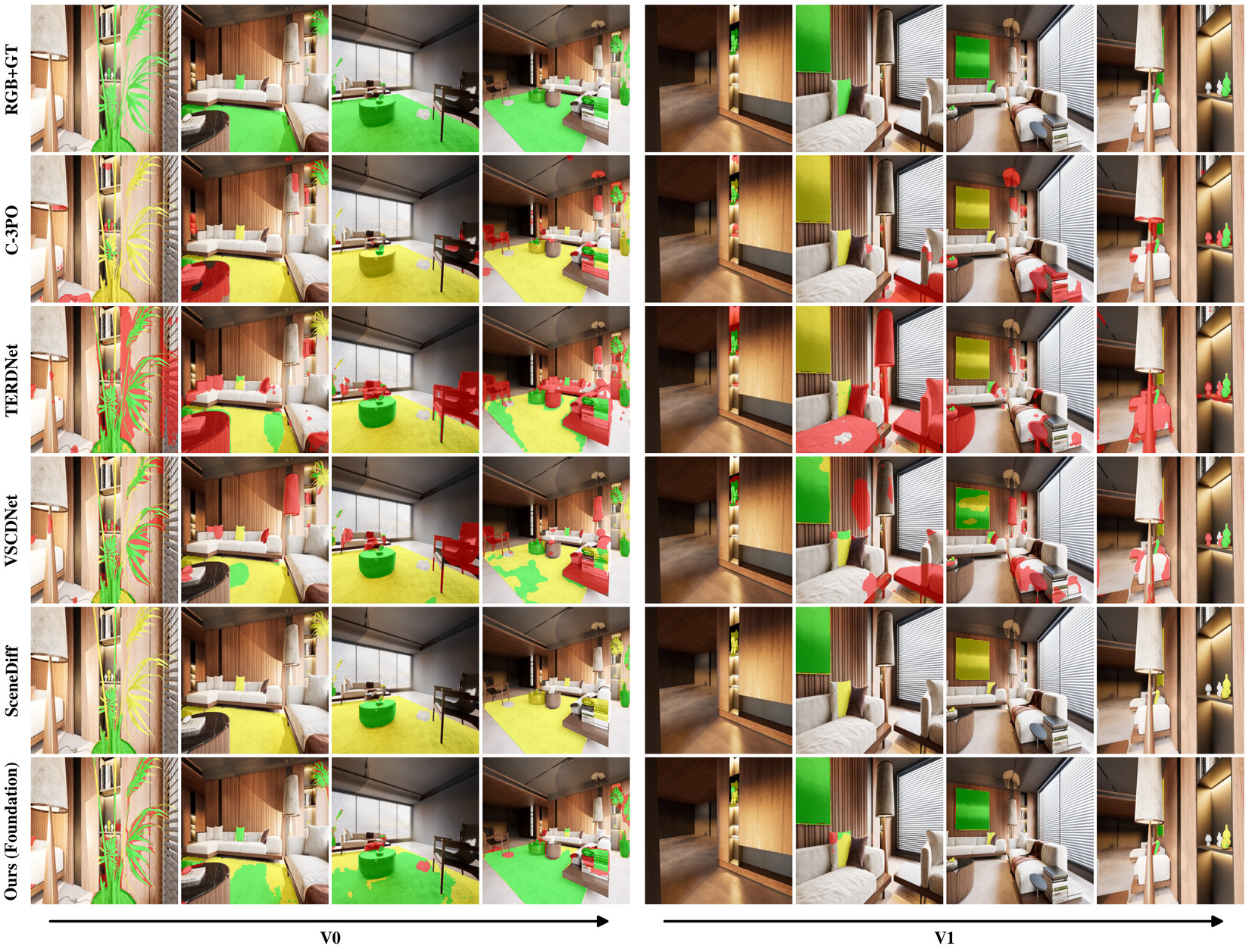}
        \caption{\textbf{Qualitative result of the VSCD dataset.} Stretched vertically to provide a better view.}
        \label{fig:vscd_dataset_result}
    \end{figure}
    
    \begin{figure}[htb]
        \centering
        \includegraphics[width=0.9\linewidth]{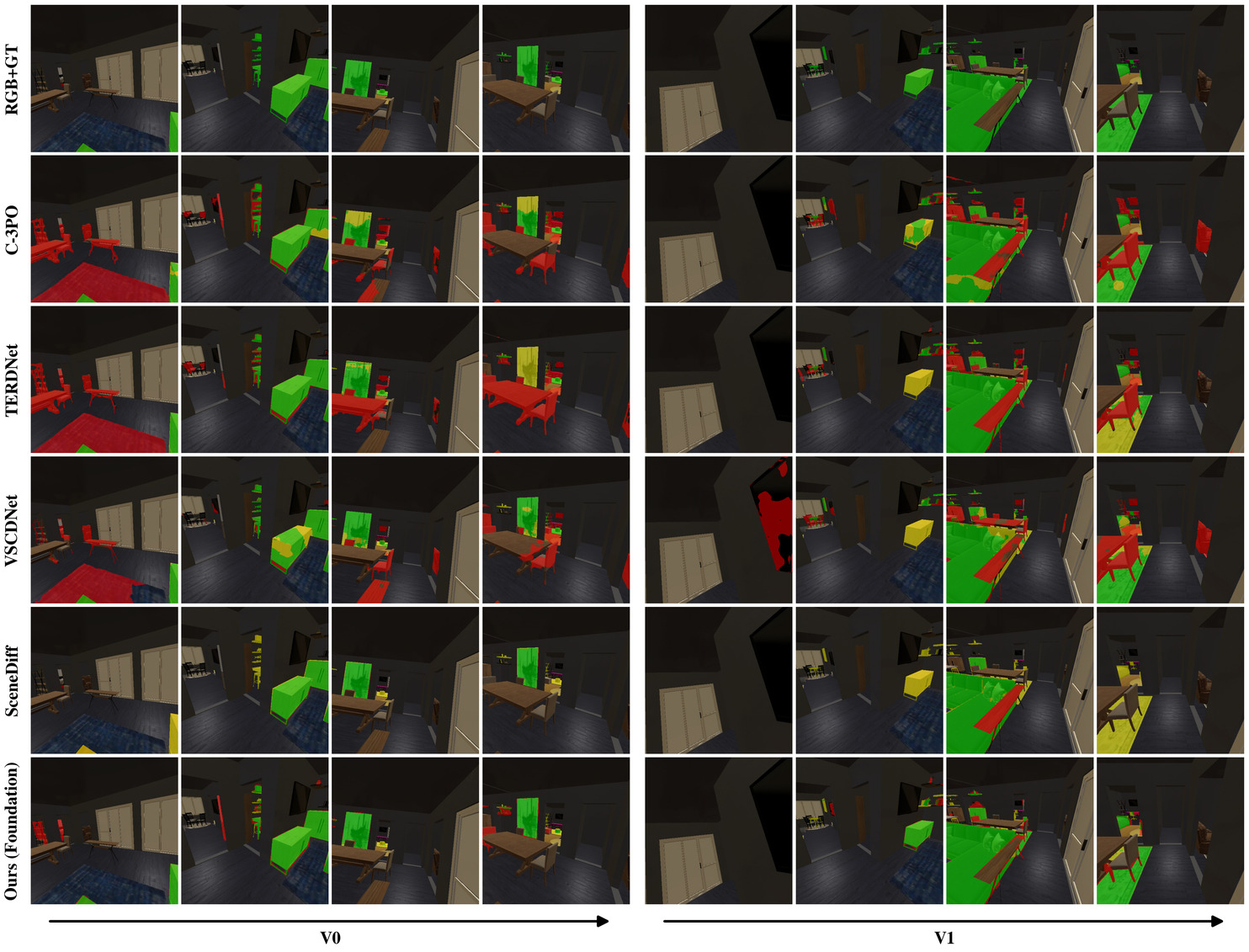}
        \caption{\textbf{Qualitative result of the \modelName{}-CD-HSSD dataset.}}
        \label{fig:hssd_cd_result}
    \end{figure}
    
    \begin{figure}[htb]
        \centering
        \includegraphics[width=0.9\linewidth]{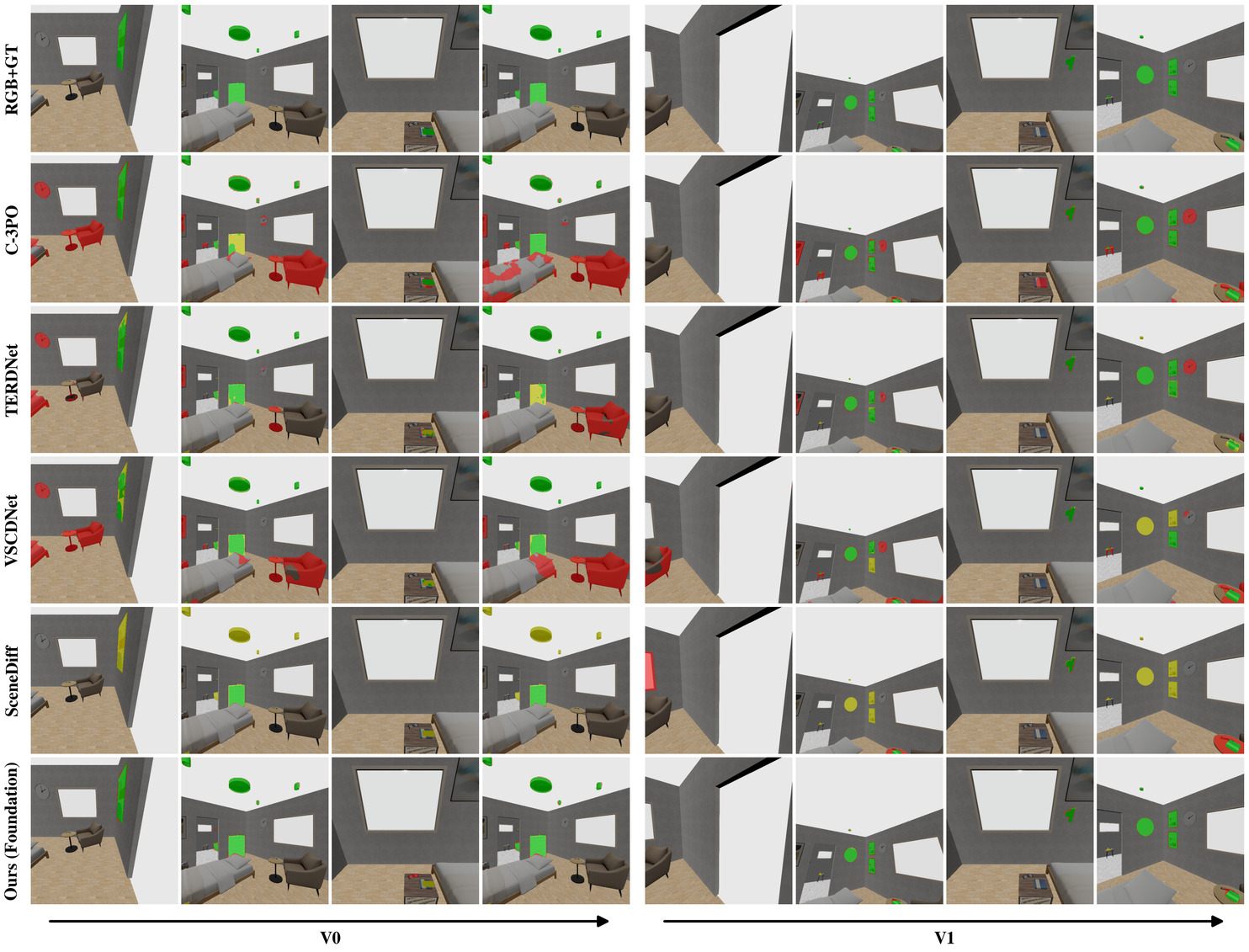}
        \caption{\textbf{Qualitative result of the \modelName{}-CD-SceneSmith dataset.}}
        \label{fig:scenesmith_result}
    \end{figure}
    
}
\newcommand{\slamSkipRateFigure}{
\begin{figure}[tbh]
    \centering
    \includegraphics[width=\linewidth]{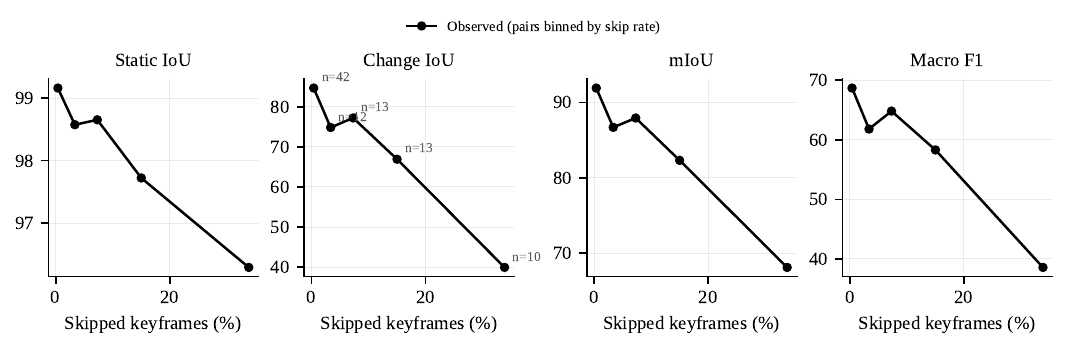}
    \caption{\textbf{Change-detection accuracy of \modelName{}-SLAM versus the keyframe skip rate.} Accuracy drops as more keyframes are skipped. Each point is one group of test pairs, plotted at its mean skip rate; $n$ is the number of pairs in the group.}
    \label{fig:slam_skip_rate}
\end{figure}
}
\newtheorem{theorem}{Theorem}
\newtheorem{problem}{Problem}
\newtheorem{trule}[theorem]{Rule}
\newtheorem{corollary}[theorem]{Corollary}
\newtheorem{conjecture}[theorem]{Conjecture}
\newtheorem{lemma}[theorem]{Lemma}
\newtheorem{assumption}[theorem]{Assumption}
\newtheorem{definition}[theorem]{Definition}
\newtheorem{proposition}[theorem]{Proposition}
\newtheorem{remark}[theorem]{Remark}
\newtheorem{example}[theorem]{Example}
\newtheorem{fact}[theorem]{Fact}

\newcommand{\cf}{\emph{cf.}\xspace}
\newcommand{\Figure}{Fig.}
\newcommand{\bdmath}{\begin{dmath}}
\newcommand{\edmath}{\end{dmath}}
\newcommand{\beq}{\begin{equation}}
\newcommand{\eeq}{\end{equation}}
\newcommand{\bdm}{\begin{displaymath}}
\newcommand{\edm}{\end{displaymath}}
\newcommand{\bea}{\begin{eqnarray}}
\newcommand{\eea}{\end{eqnarray}}
\newcommand{\beal}{\beq \begin{array}{ll}}
\newcommand{\eeal}{\end{array} \eeq}
\newcommand{\beas}{\begin{eqnarray*}}
\newcommand{\eeas}{\end{eqnarray*}}
\newcommand{\ba}{\begin{array}}
\newcommand{\ea}{\end{array}}
\newcommand{\bit}{\begin{itemize}}
\newcommand{\eit}{\end{itemize}}
\newcommand{\ben}{\begin{enumerate}}
\newcommand{\een}{\end{enumerate}}
\newcommand{\expl}[1]{&&\qquad\text{\color{gray}(#1)}\nonumber}

\newcommand{\calA}{{\cal A}}
\newcommand{\calB}{{\cal B}}
\newcommand{\calC}{{\cal C}}
\newcommand{\calD}{{\cal D}}
\newcommand{\calE}{{\cal E}}
\newcommand{\calF}{{\cal F}}
\newcommand{\calG}{{\cal G}}
\newcommand{\calH}{{\cal H}}
\newcommand{\calI}{{\cal I}}
\newcommand{\calJ}{{\cal J}}
\newcommand{\calK}{{\cal K}}
\newcommand{\calL}{{\cal L}}
\newcommand{\calM}{{\cal M}}
\newcommand{\calN}{{\cal N}}
\newcommand{\calO}{{\cal O}}
\newcommand{\calP}{{\cal P}}
\newcommand{\calQ}{{\cal Q}}
\newcommand{\calR}{{\cal R}}
\newcommand{\calS}{{\cal S}}
\newcommand{\calT}{{\cal T}}
\newcommand{\calU}{{\cal U}}
\newcommand{\calV}{{\cal V}}
\newcommand{\calW}{{\cal W}}
\newcommand{\calX}{{\cal X}}
\newcommand{\calY}{{\cal Y}}
\newcommand{\calZ}{{\cal Z}}

\newcommand{\setA}{\textsf{A}}
\newcommand{\setB}{\textsf{B}}
\newcommand{\setC}{\textsf{C}}
\newcommand{\setD}{\textsf{D}}
\newcommand{\setE}{\textsf{E}}
\newcommand{\setF}{\textsf{F}}
\newcommand{\setG}{\textsf{G}}
\newcommand{\setH}{\textsf{H}}
\newcommand{\setI}{\textsf{I}}
\newcommand{\setJ}{\textsf{J}}
\newcommand{\setK}{\textsf{K}}
\newcommand{\setL}{\textsf{L}}
\newcommand{\setM}{\textsf{M}}
\newcommand{\setN}{\textsf{N}}
\newcommand{\setO}{\textsf{O}}
\newcommand{\setP}{\textsf{P}}
\newcommand{\setQ}{\textsf{Q}}
\newcommand{\setR}{\textsf{R}}
\newcommand{\setS}{\textsf{S}}
\newcommand{\setT}{\textsf{T}}
\newcommand{\setU}{\textsf{U}}
\newcommand{\setV}{\textsf{V}}
\newcommand{\setX}{\textsf{X}}
\newcommand{\setY}{\textsf{Y}}
\newcommand{\setZ}{\textsf{Z}}

\newcommand{\smallheading}[1]{\textit{#1}: }
\newcommand{\algostep}[1]{{\small\texttt{#1:}}\xspace}
\newcommand{\etal}{\emph{et~al.}\xspace}
\newcommand{\setal}{~\emph{et~al.}\xspace}
\newcommand{\eg}{\emph{e.g.,}\xspace}
\newcommand{\ie}{\emph{i.e.,}\xspace}
\newcommand{\myParagraph}[1]{{\bf #1.}\xspace}
\newcommand{\M}[1]{{\bm #1}} %
\renewcommand{\boldsymbol}[1]{{\bm #1}}
\newcommand{\algoname}[1]{\textsc{#1}} %

\newcommand{\towrite}{{\color{red} To write}\xspace}
\newcommand{\xxx}{{\color{red} XXX}\xspace}
\newcommand{\AC}[1]{{\color{red} \textbf{AC}: #1}}
\newcommand{\LC}[1]{{\color{red} \textbf{LC}: #1}}
\newcommand{\FM}[1]{{\color{red} \textbf{FM}: #1}}
\newcommand{\RT}[1]{{\color{blue} \textbf{RT}: #1}}
\newcommand{\VT}[1]{{\color{blue} \textbf{VT}: #1}}
\newcommand{\hide}[1]{}
\newcommand{\wrt}{w.r.t.\xspace}
\newcommand{\highlight}[1]{{\color{red} #1}}
\newcommand{\tocheck}[1]{{\color{brown} #1}}
\newcommand{\grayout}[1]{{\color{gray} #1}}
\newcommand{\grayText}[1]{{\color{gray} \text{#1} }}
\newcommand{\grayMath}[1]{{\color{gray} #1 }}
\newcommand{\hiddenText}{{\color{gray} hidden text.}}
\newcommand{\hideWithText}[1]{\hiddenText}
\newcommand{\marginNote}[1]{\marginpar{\small \red{#1}}}

\newcommand{\NA}{{\sf n/a}}
\newcommand{\versus}{\scenario{VS}\xspace}

\newcommand{\kron}{\otimes}
\newcommand{\dist}{\mathbf{dist}}
\newcommand{\iter}{\! \rm{iter.} \;}
\newcommand{\leqt}{\!\!\! < \!\!\!}
\newcommand{\geqt}{\!\!\! > \!\!\!}
\newcommand{\mysetminus}{-} %
\newcommand{\powerset}{\mathcal{P}}
\newcommand{\Int}[1]{ { {\mathbb Z}^{#1} } }
\newcommand{\Natural}[1]{ { {\mathbb N}^{#1} } }
\newcommand{\Complex}[1]{ { {\mathbb C}^{#1} } }
\newcommand{\one}{ {\mathbf{1}} }
\newcommand{\subject}{\text{ subject to }}
\newcommand{\normsq}[2]{\left\|#1\right\|^2_{#2}}
\newcommand{\norm}[1]{\left\| #1 \right\|}
\newcommand{\normsqs}[2]{\|#1\|^2_{#2}}
\newcommand{\infnorm}[1]{\|#1\|_{\infty}}
\newcommand{\zeronorm}[1]{\|#1\|_{0}}
\newcommand{\onenorm}[1]{\|#1\|_{1}}
\newcommand{\lzero}{\ell_{0}}
\newcommand{\lone}{\ell_{1}}
\newcommand{\linf}{\ell_{\infty}}

\newcommand{\EV}{\mathbb{E}}
\newcommand{\erf}{{\mathbf{erf}}}
\newcommand{\prob}[1]{{\mathbb P}\left(#1\right)}
\newcommand{\tran}{^{\mathsf{T}}}
\newcommand{\traninv}{^{-\mathsf{T}}}
\newcommand{\diag}[1]{\mathrm{diag}\left(#1\right)}
\newcommand{\trace}[1]{\mathrm{tr}\left(#1\right)}
\newcommand{\conv}[1]{\mathrm{conv}\left(#1\right)}
\newcommand{\polar}[1]{\mathrm{polar}\left(#1\right)}
\newcommand{\rank}[1]{\mathrm{rank}\left(#1\right)}
\newcommand{\e}{{\mathrm e}}
\newcommand{\inv}{^{-1}}
\newcommand{\pinv}{^\dag}
\newcommand{\until}[1]{\{1,\dots, #1\}}
\newcommand{\ones}{{\mathbf 1}}
\newcommand{\zero}{{\mathbf 0}}
\newcommand{\eye}{{\mathbf I}}
\newcommand{\vect}[1]{\left[\begin{array}{c}  #1  \end{array}\right]}
\newcommand{\matTwo}[1]{\left[\begin{array}{cc}  #1  \end{array}\right]}
\newcommand{\matThree}[1]{\left[\begin{array}{ccc}  #1  \end{array}\right]}
\newcommand{\dss}{\displaystyle}
\newcommand{\Real}[1]{ { {\mathbb R}^{#1} } }
\newcommand{\reals}{\Real{}}
\newcommand{\opt}{^{\star}}
\newcommand{\only}{^{\alpha}}
\newcommand{\copt}{^{\text{c}\star}}
\newcommand{\atk}{^{(k)}}
\newcommand{\att}{^{(t)}}
\newcommand{\at}[1]{^{(#1)}}
\newcommand{\attau}{^{(\tau)}}
\newcommand{\atc}[1]{^{(#1)}}
\newcommand{\atK}{^{(K)}}
\newcommand{\atj}{^{(j)}}
\newcommand{\projector}{{\tt projector}}
\newcommand{\setdef}[2]{ \{#1 \; {:} \; #2 \} }
\newcommand{\smalleye}{\left(\begin{smallmatrix}1&0\\0&1\end{smallmatrix}\right)}

\newcommand{\SEtwo}{\ensuremath{\mathrm{SE}(2)}\xspace}
\newcommand{\SE}[1]{\ensuremath{\mathrm{SE}(#1)}\xspace}
\newcommand{\SEthree}{\ensuremath{\mathrm{SE}(3)}\xspace}
\newcommand{\SOtwo}{\ensuremath{\mathrm{SO}(2)}\xspace}
\newcommand{\SOthree}{\ensuremath{\mathrm{SO}(3)}\xspace}
\newcommand{\Othree}{\ensuremath{\mathrm{O}(3)}\xspace}
\newcommand{\SOn}{\ensuremath{\mathrm{SO}(n)}\xspace}
\newcommand{\SO}[1]{\ensuremath{\mathrm{SO}(#1)}\xspace}
\newcommand{\On}{\ensuremath{\mathrm{O}(n)}\xspace}
\newcommand{\sotwo}{\ensuremath{\mathrm{so}(2)}\xspace}
\newcommand{\sothree}{\ensuremath{\mathrm{so}(3)}\xspace}
\newcommand{\intexpmap}[1]{\mathrm{Exp}\left(#1\right)}
\newcommand{\intlogmap}[1]{\mathrm{Log}\left(#1\right)}
\newcommand{\logmapz}[1]{\mathrm{Log}_0(#1)}
\newcommand{\loglikelihood}{\mathrm{log}\calL}
\newcommand{\intprinlogmap}{\mathrm{Log}_0}
\newcommand{\expmap}[1]{\intexpmap{#1}}
\newcommand{\expmaps}[1]{\langle #1 \rangle_{2\pi}}
\newcommand{\biggexpmap}[1]{\left\langle #1 \right\rangle_{2\pi}}
\newcommand{\logmap}[1]{\intlogmap{#1}}
\newcommand{\round}[1]{\mathrm{round}\left( #1 \right)}

\newcommand{\MA}{\M{A}}
\newcommand{\MB}{\M{B}}
\newcommand{\MC}{\M{C}}
\newcommand{\MD}{\M{D}}
\newcommand{\ME}{\M{E}}
\newcommand{\MJ}{\M{J}}
\newcommand{\MK}{\M{K}}
\newcommand{\MG}{\M{G}}
\newcommand{\MM}{\M{M}}
\newcommand{\MN}{\M{N}}
\newcommand{\MP}{\M{P}}
\newcommand{\MQ}{\M{Q}}
\newcommand{\MU}{\M{U}}
\newcommand{\MR}{\M{R}}
\newcommand{\MS}{\M{S}}
\newcommand{\MI}{\M{I}}
\newcommand{\MV}{\M{V}}
\newcommand{\MF}{\M{F}}
\newcommand{\MH}{\M{H}}
\newcommand{\ML}{\M{L}}
\newcommand{\MO}{\M{O}}
\newcommand{\MT}{\M{T}}
\newcommand{\MX}{\M{X}}
\newcommand{\MY}{\M{Y}}
\newcommand{\MW}{\M{W}}
\newcommand{\MZ}{\M{Z}}
\newcommand{\MSigma}{\M{\Sigma}} 
\newcommand{\MOmega}{\M{\Omega}}
\newcommand{\MPhi}{\M{\Phi}}
\newcommand{\MPsi}{\M{\Psi}}
\newcommand{\MDelta}{\M{\Delta}}
\newcommand{\MLambda}{\M{\Lambda}}

\newcommand{\Rtheta}{\boldsymbol{R}}
\newcommand{\symf}{f} %

\newcommand{\angledomain}{(-\pi,+\pi]}

\newcommand{\MCB}{\mathsf{MCB}}
\newcommand{\FCM}{\mathsf{FCM}}
\newcommand{\FCB}{\mathsf{FCB}}
\newcommand{\cyclemap}[1]{\calC^{\calG}\left(#1\right)}
\newcommand{\incidencemap}[1]{\calA^{\calG}\left(#1\right)}
\newcommand{\cyclemapk}{\calC^{\calG}_{k}}
\newcommand{\incidencemapij}{\calA^{\calG}_{ij}}
\renewcommand{\ij}{_{ij}}
\newcommand{\foralledges}{\forall(i,j) \in \calE}
\newcommand{\sumalledges}{
     \displaystyle
     \sum_{(i,j) \in \calE}}
\newcommand{\sumalledgesm}{
     \displaystyle
     \sum_{i=1}^{m}}
\newcommand{\T}{\mathsf{T}}
\newcommand{\To}{\T_{\rm o}}
\newcommand{\Tm}{\T_{\rm m}}
\newcommand{\MCBa}{\MCB_{\mathsf{a}}}
\newcommand{\FCBo}{\FCB_{\mathsf{o}}}
\newcommand{\FCBm}{\FCB_{\mathsf{m}}}

\newcommand{\algoonlyname}{MOLE2D}
\newcommand{\algoml}{{\smaller\sf \algoonlyname}\xspace}
\newcommand{\algocyclebasis}{\algoname{compute-cycle-basis}}
\newcommand{\scenario}[1]{{\smaller \sf#1}\xspace}
\newcommand{\toro}{{\smaller\sf Toro}\xspace}
\newcommand{\gtwoo}{{\smaller\sf g2o}\xspace}
\newcommand{\gtwooST}{{\smaller\sf g2oST}\xspace}
\newcommand{\gtwood}{{\smaller\sf g2o{10}}\xspace}
\newcommand{\gtsam}{{\smaller\sf gtsam}\xspace}
\newcommand{\isam}{{\smaller\sf iSAM}\xspace}
\newcommand{\lago}{{\smaller\sf LAGO}\xspace}
\newcommand{\egtwoo}{{\smaller\sf \algoonlyname+g2o}\xspace}

\newcommand{\rim}{\scenario{rim}}
\newcommand{\cubicle}{\scenario{cubicle}}
\newcommand{\sphere}{\scenario{sphere}}
\newcommand{\sphereHard}{\scenario{sphere-a}}
\newcommand{\garage}{\scenario{garage}}
\newcommand{\torus}{\scenario{torus}}
\newcommand{\oneloop}{\scenario{circle}}
\newcommand{\intel}{\scenario{INTEL}}
\newcommand{\bovisa}{\scenario{Bovisa}}
\newcommand{\bov}{\scenario{B25b}}
\newcommand{\fra}{\scenario{FR079}}
\newcommand{\frb}{\scenario{FRH}}
\newcommand{\csail}{\scenario{CSAIL}}
\newcommand{\Ma}{\scenario{M3500}}
\newcommand{\Mb}{\scenario{M10000}}
\newcommand{\ATE}{\scenario{ATE}}
\newcommand{\CVX}{\scenario{CVX}}
\newcommand{\NEOS}{\scenario{NEOS}}
\newcommand{\sdptThree}{\scenario{sdpt3}}
\newcommand{\MOSEK}{\scenario{MOSEK}}
\newcommand{\NESTA}{\scenario{NESTA}}
\newcommand{\vertigo}{\scenario{Vertigo}}
\newcommand{\SDPA}{\scenario{SDPA}}

\newcommand{\cvx}{{\sf cvx}\xspace}

\newcommand{\blue}[1]{{\color{blue}#1}}
\newcommand{\green}[1]{{\color{green}#1}}
\newcommand{\red}[1]{{\color{red}#1}}

\newcommand{\linkToPdf}[1]{\href{#1}{\blue{(pdf)}}}
\newcommand{\linkToPpt}[1]{\href{#1}{\blue{(ppt)}}}
\newcommand{\linkToCode}[1]{\href{#1}{\blue{(code)}}}
\newcommand{\linkToWeb}[1]{\href{#1}{\blue{(web)}}}
\newcommand{\linkToVideo}[1]{\href{#1}{\blue{(video)}}}
\newcommand{\linkToMedia}[1]{\href{#1}{\blue{(media)}}}
\newcommand{\award}[1]{\xspace} %

\newcommand{\vpose}{\boldsymbol{x}}
\newcommand{\vDelta}{\boldsymbol{\Delta}}
\newcommand{\vposesub}{\hat{\vpose}}
\newcommand{\vpossub}{\hat{\vpos}}
\newcommand{\vthetasub}{\hat{\vtheta}}
\newcommand{\Pthetasub}{\MP_\vtheta}
\newcommand{\thetasub}{\hat{\theta}}
\newcommand{\vposecorr}{\tilde{\vpose}}
\newcommand{\vposcorr}{\tilde{\vpos}}
\newcommand{\vthetacorr}{\tilde{\vtheta}}
\newcommand{\vposestar}{{\vpose}^{\star}}
\newcommand{\vposstar}{{\vpos}^{\star}}
\newcommand{\vthetastar}{{\vtheta}^{\star}}
\newcommand{\vcthetastar}{{\vctheta}^{\star}}
\newcommand{\pose}{\boldsymbol{x}}
\newcommand{\pos}{\boldsymbol{p}}
\newcommand{\mease}{z} %
\newcommand{\meas}{\boldsymbol{z}} %
\newcommand{\meashat}{\hat{\meas}}

\maketitle
\fancyhead{}
\renewcommand{\headrulewidth}{0pt}

\teaserFigure

\begin{abstract}
Robots operating in dynamic environments require reliable detection of how their surroundings change over time. Existing learning-based methods largely rely on pairwise 2D image features, which struggle under large viewpoint changes and occlusions, are sensitive to noise, and show limited generalization across domains, while explicit 3D approaches typically require costly offline optimization. We show that the implicit 3D knowledge of Geometric Foundation Models (GFMs) provides a strong basis for addressing these limitations. We introduce \modelName{}, which adapts GFM features for joint scene change detection and 3D reconstruction. To address data scarcity and take a step toward a foundation model for scene change detection, we introduce a large-scale benchmark comprising two synthetic datasets and one real-world dataset, and train jointly across diverse datasets to improve cross-domain generalization. We further introduce \modelName{}-SLAM, a real-time system designed for robotics, which performs online change detection and change-aware 4D mapping. Across benchmarks, our framework substantially outperforms existing baselines, with gains of up to 42.01\% in change IoU and 27.91\% in F1, while supporting scalable deployment in changing real-world environments. Code and datasets are available at the \href{https://www.multyxu.com/argos}{\textcolor{blue}{project website}}.

\end{abstract}

\section{Introduction}
Long-term autonomy requires agents to continuously update their spatial knowledge in the face of a changing world. A moved tool may invalidate a planned interaction, while newly placed furniture may block a previously traversable route. For robots to operate autonomously in such environments, they must not only perceive and understand how the environment changes over time, but also maintain an accurate memory of those changes for subsequent navigation and manipulation. A map that faithfully reflects the current state of the environment is therefore a key component of long-term robot autonomy. This motivates the need for robust change detection that can operate under large viewpoint variations, occlusions, and continuous observations during real-world robot operation.

Existing approaches have progressed from image pairs to multi-view observations, but geometric ambiguity and limited generalization remain central challenges. Learned detectors predominantly compare 2D features, making large viewpoint shifts and occlusions difficult to resolve~\citep{Wang23pr-C3PO, Lin25icra-RobustSCD, Yoon26icra-TERDNet, Yoon26icml-VSCD}. Explicit 3D pipelines provide spatial context, but often infer change through heavy optimization with handcrafted comparison rules~\citep{Wu25eccv-SceneDiff, Galappaththige26cvpr-changeInRealTime}. Meanwhile, benchmark-specific training and scarce long-sequence annotations restrict generalization across environments and observation trajectories~\citep{Kim25cvpr-GeSCF, Yoon26icml-VSCD}.  

To address these limitations, we introduce  \textbf{\modelName{}} (\textbf{\underline{A}}dapt \textbf{\underline{R}}ich \textbf{\underline{G}}eometric Priors for Generalizable \textbf{\underline{O}}nline \textbf{\underline{S}}cene-Change-Detection), a feedforward model that learns changes directly from the implicit representations of Geometric Foundation Models (GFMs)~\citep{Wang25cvpr-vggt, wang26cvpr_vggtomega}.  Our key insight is that the geometric knowledge learned to associate scene content across viewpoints also provides a prior for recognizing inconsistent content across sessions. Crucially, such inconsistency requires to be interpreted through co-visibility: observing that the space previously occupied by an object is now empty provides evidence of removal, whereas simply failing to see an occluded object does not. We realize this insight through a co-visibility-aware inter-session comparator that incorporates camera information and session membership to interpret feature differences. As illustrated in Figure~\ref{fig:teaser}, \modelName{} jointly predicts change masks and scene geometry from two unposed RGB image collections, retaining reconstruction capability while learning to compare different scene states.

To support learning beyond individual benchmarks, we introduce \modelName{}-CD, which comprises two large-scale synthetic datasets with automatically generated robot traversals and a challenging real-world evaluation set. Joint training on \modelName{}-CD and existing datasets provides diverse change supervision across domains, taking a step toward a foundation model for scene change detection. Experiments demonstrate substantial gains across paired-image and video benchmarks; even with synthetic-only change supervision, \modelName{} surpasses learned baselines fine-tuned on real data on real-world evaluations. We further integrate \modelName{} with VGGT-SLAM~\citep{Maggio26rss-VGGT-SLAM2} into \modelName{}-SLAM. This system supports online reconstruction and change updates for change-aware 4D mapping.

Our contributions are summarized as follows:
\begin{itemize}
    \item \textbf{Geometrically grounded change detection.} We introduce \modelName{}, a model that adapts implicit 3D priors through co-visibility-aware inter-session comparison for joint change detection and reconstruction from a pair of RGB sequences.
    \item \textbf{Large-scale dataset and generalization.} We introduce the \modelName{}-CD dataset and demonstrate that leveraging pretrained GFM features together with multi-dataset training enables strong benchmark performance and promising real-world generalization.
    \item \textbf{Online change-aware mapping.} We develop \modelName{}-SLAM, which connects learned change predictions to online reconstruction of a change-aware point-cloud map.
\end{itemize}

\section{Related Work}
\noindent\textbf{Geometric Foundation Model.}
Multi-view geometry and SLAM establish the spatial relationships needed to reconstruct scenes from visual observations~\citep{hartley03book-mvgcv, Cadena16tro-SLAMsurvey, zhou2023dynpoint}. Geometric Foundation Models (GFMs), including VGGT, VGGT-$\Omega$, and $\pi^3$ predict camera poses and dense geometry through feedforward inference~\citep{Wang25cvpr-vggt, wang26cvpr_vggtomega, wang26iclr-pi3}. D4RT and PAGE-4D extend feedforward geometric perception to dynamic scenes~\citep{zhang26cvpr-d4rt, zhou26icml_PAGE4D}, while WT-DiT connects spatial reconstruction with generative world modeling~\citep{zhang26arxic-wt-dit}. 
Beyond reconstruction, \citet{Bratulić26arsiv-geometricunderstandinglearnedpriors} probe VGGT’s internal representations to investigate the geometric information encoded by GFMs, revealing that these representations implicitly capture epipolar geometry.
We study how these reconstruction-trained features can support co-visibility-aware comparison of different scene states across sessions. Rather than using only reconstruction outputs, our approach uses GFM's latent representations and tokens to inform change detection.

\noindent\textbf{Change Detection.}
Scene change detection (SCD) traditionally localizes changes between image pairs from similar viewpoints~\citep{Sakurada15bmvc-PCD, alcantarilla18ar-VLCMUCD, Sakurada20icra-CSCDNet&PSCD, Park21iros-ChangeSim}, using learned pairwise features, temporal interaction, segmentation, or cross-view matching~\citep{Varghese18eccv_ChangeNet, Daudt18icip-FCSiam, Chen21iv-DR-TANet, Wang23pr-C3PO, Sachdeva23wacv_CYWS, Sachdeva23iccvw_CYWS-3D}. Recent methods leverage pretrained representations such as DINOv2~\citep{Oquab23arxiv-dinov2} and SAM~\citep{Kirillov23iccv-SegmentAnything} through task-specific adaptation~\citep{Lin25icra-RobustSCD, Yoon26icra-TERDNet} or zero-shot comparison~\citep{Cho25aaai-ZSSCD, Kannan25-ZeroSCD, Kim25cvpr-GeSCF}. Multi-view work extends SCD to unaligned observations~\citep{Wald19iccv-RIO, Wu25eccv-SceneDiff, Galappaththige25cvpr-MV3DCD}: VSCD~\citep{Yoon26icml-VSCD} learns multi-reference cross-frame correspondence, SceneDiff~\citep{Wu25eccv-SceneDiff} selects co-visible views using reconstructed geometry, and O-SCD~\citep{Galappaththige26cvpr-changeInRealTime} performs self-supervised fusion over an offline Gaussian Splatting reconstruction~\citep{Kerbl23Ttog-GaussianSplatting}. However, learned methods can still struggle with cross-domain transfer, while zero-shot approaches often require costly reconstruction or optimization. In contrast, \modelName{} performs cross-session comparison directly in the latent space of a frozen GFM and improves transfer through joint training across diverse datasets.

\noindent\textbf{Robot Mapping in Dynamic Environments.}
The literature on long-term mapping and SLAM has investigated robot operation in changing environments~\citep{Fehr17icra-tsdf-cd, fu22ral-PlaneSDF-cd, Schmid22icra-Panoptic-mapping}. Khronos~\citep{Schmid24rss-khronos} constructs dense spatio-temporal maps, identifying quasi-static changes through global optimization over TSDF-based maps~\citep{curless96acm-tsdf}. SuperMap~\citep{zhao26rss-supermap} extends this direction with a real-time online system and open-set semantic capabilities using vision foundation models~\citep{Oquab23arxiv-dinov2, ravi25iclr-sam2}. For LiDAR, Chamelion~\citep{Jang26ral-Chamelion} uses a dual-head network for change classification and long-term map maintenance, while ELite~\citep{Gil25ral-ELite} models ephemerality for lifelong mapping. These systems typically detect change through explicit map geometry or object associations rather than leveraging advances in learned change detection. We bridge the two with \modelName{}-SLAM, which combines online reconstruction and change detection to maintain a change-aware point-cloud map: SLAM supplies the spatio-temporal structure for long-term deployment, while learned RGB change detection improves change identification beyond geometry alone.

\section{Problem Formulation}
\label{sec:problem_formulation}

We consider a robot that revisits a scene and needs to detect changes while reconstructing the scene.

\textbf{Two-visit Change Detection.}
The inputs are two sets of \emph{unposed} RGB images from two visits, $\mathbb{I}^{v_0}=\{I_i^{v_0}\}_{i=1}^{K}$ and $\mathbb{I}^{v_1}=\{I_j^{v_1}\}_{j=1}^{N}$. The visits may differ in length, trajectory, viewpoint, and overlap. The goal is to predict a binary change mask for every image,
\begin{equation}
\mathbb{M}^{v_0}=\{M_i^{v_0}\}_{i=1}^{K}, \qquad
\mathbb{M}^{v_1}=\{M_j^{v_1}\}_{j=1}^{N}, \qquad
M\in\{0,1\}^{H\times W}.
\end{equation}
Changed content is labeled in the visit where it is physically present. For instance, if a new object becomes visible in $I_j^{v_1}$, the corresponding pixels are labeled in $M_j^{v_1}$. Conversely, if an object visible in $I_i^{v_0}$ is absent from a co-visible frame $I_j^{v_1}$, its pixels are labeled in $M_i^{v_0}$.

\textbf{Change-aware Point-cloud Map.}
Naively reconstructing the scene from both visits produces a joint point cloud $\mathbb{P}$ that mixes the static background with changed elements observed across visits. Instead, we seek a change-aware point-cloud map $\mathbb{P}(v)$, where $v\in\{v_0,v_1\}$ specifies the queried visit, that disentangles these changes and faithfully represents the corresponding scene state. Specifically, $\mathbb{P}(v_0)$ removes geometry corresponding to elements present only in $v_1$, while retaining the shared static background and visit-specific geometry from $v_0$; $\mathbb{P}(v_1)$ is defined symmetrically. For example, if a cup in $v_0$ is replaced by a bowl in $v_1$, the joint reconstruction may contain overlapping geometry from both objects. Querying $v_0$ removes the bowl geometry and recovers the scene with the cup.

\section{Method}
\label{sec:method}
\LayerAttentionViz

This section describes \modelName. We first motivate our design, then describe the architecture, training data, and finally integration with a state-of-the-art SLAM system to form \modelName-SLAM. 

\subsection{Motivation}
\label{sec:method_motivation}

Learning-based scene change detection often compares 2D image features, making it difficult to distinguish true changes from viewpoint- or occlusion-induced appearance differences. We instead view change detection as inherently 3D: evidence for change requires both observing the same scene region across visits and finding inconsistent content there. 

GFMs provide a natural representation for this reasoning. In particular,  VGGT-$\Omega$~\citep{wang26cvpr_vggtomega} is pretrained with a matching loss $\mathcal{L}_{\mathrm{match}}$ that encourages features corresponding to the same 3D location to align across views. Inspired by the attention analysis used in PAGE-4D~\citep{zhou26icml_PAGE4D} and VGGT-SLAM 2.0~\citep{Maggio26rss-VGGT-SLAM2}, we similarly probe pretrained VGGT-$\Omega$ across visits. Figure~\ref{fig:layer_attention_viz} shows that its intermediate features respond strongly to corresponding object locations across viewpoints, but weakly when the object is absent, suggesting useful cues for co-visibility and cross-visit consistency. Guided by this observation, \modelName{} uses pose and session information to adapt these implicit geometric features for co-visibility-aware cross-session comparison, jointly predicting scene changes, camera poses, and depth. Additional attention analyses are presented in Appendix~\ref{app:backbone_attention}.

\subsection{Model Architecture}

Illustrated in Figure~\ref{fig:architecture}, \modelName{} is a unified model that jointly predicts change masks and scene geometry from the two visits. To enable cross-visit change reasoning, it augments VGGT-$\Omega$ with a \emph{Co-visibility-aware Spatio-Temporal Comparator} comprising a Camera Reader, Inter-session Cross-Attention, and a DPT Fusion Head. We index the combined $n=K+N$ images by $r \in \{1,\ldots,n\}$, with the first $K$ images from $v_0$ and the remaining $N$ from $v_1$. The model implements
\begin{equation}
f_\theta\big(\mathbb{I}^{v_0},\mathbb{I}^{v_1}\big)
= \big\{\hat{M}_r,\hat{D}_r,\hat{g}_r\big\}_{r=1}^{n},
\end{equation}
where $\hat{M}_r \in \{0,1\}^{H\times W}$ is a change mask, $\hat{D}_r \in \mathbb{R}^{H\times W}$ is a depth map, and $\hat{g}_r \in \mathbb{R}^{9}$ encodes camera extrinsics and intrinsics. Unprojecting the predicted depth using the estimated cameras and disentangle changed regions with the masks yields a change-aware point-cloud map $\mathbb{P}(v)$.

\textbf{Backbone and Tokens.} The frozen VGGT-$\Omega$ backbone~\citep{wang26cvpr_vggtomega} processes both visits jointly. We extract its features at layers $\mathcal{L}=\{4,11,17,23\}$ and denote the patch tokens for image $r$ at layer $\ell$ by $X_r^\ell \in \mathbb{R}^{T\times d}$, where $T$ is the number of image patches and $d=2048$ is the feature width. The pretrained camera and depth heads read the original backbone features; the following modules adapt them for the change branch. 
\architectureFigure

\textbf{Camera Reader.} To make patch tokens co-visibility aware, inspired by DPT's readout operator~\citep{Ranftl21iccv-DPT}, Camera Reader injects final camera features into patch tokens after selected VGGT-$\Omega$ intermediate layers. Let $c_r \in \mathbb{R}^{d}$ denote the normalized camera-head token for image $r$, immediately before it is decoded into $\hat{g}_r$. For each patch $q \in \{1,\ldots,T\}$ and layer $\ell \in \mathcal{L}$, we compute
\begin{equation}
\tilde{x}_{r,q}^{\ell}
= x_{r,q}^{\ell} + \mathrm{MLP}_{\ell}\big([c_r\,;\,x_{r,q}^{\ell}]\big),
\end{equation}
where $x_{r,q}^{\ell}$ is the $q$-th row of $X_r^\ell$, $[\cdot\,;\cdot]$ denotes channel-wise concatenation, and $\mathrm{MLP}_{\ell}:\mathbb{R}^{2d}\rightarrow\mathbb{R}^{d}$ is learned separately for each layer. The same $c_r$ is broadcast to every patch of image $r$ at all selected layers. MLP implementation details are given in Appendix~\ref{app:model_implementation}.

\textbf{Inter-session Cross-Attention.} The backbone mixes information across all images without explicitly distinguishing the two visits. We therefore use visit membership to compare their features at each layer $\ell \in \mathcal{L}$. Let $Z_{v_0}^{\ell}$ and $Z_{v_1}^{\ell}$ stack the camera, register, and camera-aware patch tokens from their respective visits. Each visit uses its own tokens as queries and the other visit's tokens as context, giving the residual updates
\begin{equation}
\bar{Z}_{v_0}^{\ell} = Z_{v_0}^{\ell} + \mathrm{CrossAttn}_{\ell}\big(Z_{v_0}^{\ell},Z_{v_1}^{\ell}\big), \qquad
\bar{Z}_{v_1}^{\ell} = Z_{v_1}^{\ell} + \mathrm{CrossAttn}_{\ell}\big(Z_{v_1}^{\ell},Z_{v_0}^{\ell}\big),
\end{equation}
where $\mathrm{CrossAttn}_{\ell}(A,B)$ denotes multi-head attention with queries from $A$ and keys and values from $B$, following the standard attention formulation~\citep{Vaswani17neurips-transformer}.
A feed-forward sublayer then maps each $\bar{Z}_{v}^{\ell}$ to $\tilde{Z}_{v}^{\ell}$, the features passed to the change head. Since change signal is symmetric, the two directions share attention and feed-forward weights, while each feature layer has its own block. The full block equations are given in Appendix~\ref{app:model_implementation}.

Intuitively, camera-augmented attention compares both co-visibility and visual appearance.
Together, these learned interactions connect the co-visibility and visual inconsistency intuition in Section~\ref{sec:method_motivation}, enabling the change head to interpret appearance differences in their geometric context.

\textbf{DPT Fusion Head.} We extract the updated patch tokens $Y_r^\ell \in \mathbb{R}^{T\times d}$ for image $r$ from $\tilde{Z}_v^\ell$ and decode them with a DPT-style head $\mathcal{H}_{\mathrm{CD}}$~\citep{Ranftl21iccv-DPT}, following VGGT-$\Omega$'s depth decoder. The head reshapes the tokens into spatial feature maps, resizes and fuses them across layers, and predicts per-pixel logits $z_r \in \mathbb{R}^{H\times W}$:
\begin{equation}
z_r = \mathcal{H}_{\mathrm{CD}}\big(\{Y_r^\ell\}_{\ell\in\mathcal{L}}\big), \qquad
\hat{p}_r = \sigma(z_r), \qquad
\hat{M}_r = \mathbf{1}[\hat{p}_r > 0.5],
\end{equation}
where $\sigma$ is the sigmoid function and the threshold is applied independently to each pixel.

\subsection{New Dataset}

Video change-detection data remain scarce for long robot traversals, hindering large-scale training. Therefore, we introduce \textbf{\modelName{}-CD}, a large-scale benchmark comprising two synthetic datasets, \textbf{\modelName{}-CD-HSSD} and \textbf{\modelName{}-CD-SceneSmith}, leveraging simulation assets from HSSD~\citep{Khanna24cvpr-HSSD} and SceneSmith~\citep{pfaff26icml-scenesmith}, and a real-world evaluation set \textbf{\modelName{}-CD-real}. Our synthetic datasets share an automated pipeline that samples scene states, plans independent trajectories to cover scene objects and navigable areas, and generates change masks from object identities. Table~\ref{tab:dataset} shows that \modelName{}-CD-HSSD covers $4\times$ more visually distinct places on average and up to $2\times$ longer maximum video duration than VSCD and SceneDiff. Furthermore, \textbf{\modelName{}-CD-real} captures challenging real-world scenarios, including cluttered workshops and opposite-direction camera traversals. See Appendix~\ref{app:dataset_generation} for additional details.

\datasetComparison

\subsection{Training}
\label{sec:training}

\textbf{Training Datasets.} We train on VL-CMU-CD~\citep{alcantarilla18ar-VLCMUCD}, PSCD~\citep{Sakurada20icra-CSCDNet&PSCD}, and ChangeSim~\citep{Park21iros-ChangeSim} from the SCD task, alongside VSCD~\citep{Yoon26icml-VSCD}, \modelName{}-CD-HSSD, \modelName{}-CD-SceneSmith, and SceneDiff~\citep{Wu25eccv-SceneDiff} from the VSCD task. We evaluate on the held-out test sets of each benchmark, as well as on \modelName{}-CD-real for additional out-of-distribution real-world evaluation.

\textbf{Loss Function and Training Details.} To handle class imbalance between sparse change pixels and dominant static regions, we apply weighted binary cross-entropy exclusively to valid (labeled) pixels. We freeze the original VGGT-$\Omega$ components and train only the three modules constituting the Co-visibility-aware Spatio-Temporal Comparator. Furthermore, we extend VGGT's training recipe by interleaving SCD and VSCD datasets across input sequence lengths ranging from 2 to 64 frames. See Appendix~\ref{app:model_implementation} for additional implementation details.

\subsection{Online Robot Mapping System}
\label{sec:online}

\textbf{Online Reconstruction, Asynchronous Change-aware Map Updates.}
We introduce \modelName{}-SLAM, an extension of VGGT-SLAM 2.0~\citep{Maggio26rss-VGGT-SLAM2} for online change detection and change-aware 4D mapping. The same \modelName{} network serves as both (i) the \emph{geometry} backbone for submap reconstruction and (ii) the \emph{change} model for cross-visit comparison. Given a prior reconstruction from the first visit $v_0$, the system incrementally reconstructs the scene during the second visit $v_1$ and aligns the incoming geometry with $v_0$ in a shared coordinate frame. Each new submap is integrated into the global reconstruction and triggers an asynchronous change-detection job. Once completed, the detected changes are incorporated into the change-aware point-cloud map $\mathbb{P}(v)$, which therefore reflects all geometry and changes accumulated up to that submap. To maintain online operation, each job retrieves only the most relevant prior keyframes rather than comparing against the entire prior visit. An efficient retrieval cascade combines cached SALAD descriptor matching, strided geometric reprojection filtering, and scene-centroid coverage search, while capping the combined input at $64$ images for sub-second inference. Change-detection jobs run asynchronously alongside reconstruction, allowing the change-aware map to be updated incrementally as new submaps arrive. Section~\ref{sec:real_world_exp} reports the runtime during real-world robot deployment, while Appendix~\ref{app:online_system} provides additional implementation details.

\section{Experiments}
\subsection{Experimental Setup}

\textbf{Baselines and Benchmark.}
We evaluate \modelName{} on the real and synthetic SCD and VSCD datasets described in Section~\ref{sec:training}. We compare against learned SCD methods (FC-EF, FC-Siam-diff, FC-Siam-conc~\citep{Daudt18icip-FCSiam}, CSCDNet~\citep{Sakurada20icra-CSCDNet&PSCD}, DR-TANet~\citep{Chen21iv-DR-TANet}, C-3PO~\citep{Wang23pr-C3PO}, RobustSCD~\citep{Lin25icra-RobustSCD}, TERDNet~\citep{Yoon26icra-TERDNet}), learned VSCD (VSCDNet~\citep{Yoon26icml-VSCD}), zero-shot SCD (ZSSCD~\citep{Cho25aaai-ZSSCD}, GeSCF~\citep{Kim25cvpr-GeSCF}) and VSCD (SceneDiff~\citep{Wu25eccv-SceneDiff}) methods. We additionally evaluate the zero-shot Gaussian-splatting method O-SCD~\citep{Galappaththige26cvpr-changeInRealTime}; due to frequent failures, its results are deferred to Appendix~\ref{app:additional_experiments}. Appendix~\ref{app:evaluation_protocol} shows more details.

\textbf{Model Variants.}
We consider four variants: \textbf{Ours}, trained specifically on each evaluation dataset; \textbf{Ours-synthetic-only}, trained exclusively on synthetic data; \textbf{Ours-finetune}, obtained by fine-tuning the synthetic variant on real-world data; and \textbf{Ours-foundation}, trained on all available datasets. Notably, \textbf{Ours-finetune} and \textbf{Ours-foundation} use the same training datasets but differ in their training strategies. Specifically, \textbf{Ours-finetune} is first trained on synthetic datasets and then fine-tuned on real-world datasets, whereas \textbf{Ours-foundation} is trained on all datasets simultaneously.

\textbf{Metrics.}
We treat change detection as binary segmentation and report static IoU, change IoU, their mean (mIoU), and change-class macro F1, all as percentages.

\subsection{Real-World Generalization}
\mainTable
\QualitativeFigure
Table~\ref{tab:main_table} shows that \textbf{Ours-synthetic-only} substantially outperforms the best fine-tuned learned baselines in change IoU (39.84 vs. 20.70 on SceneDiff and 31.18 vs. 13.50 on \modelName{}-CD-real), surpasses the zero-shot SceneDiff baseline on the SceneDiff dataset (39.84 vs. 30.31), and remains close to it on \modelName{}-CD-real (31.18 vs. 32.92). We attribute this sim-to-real generalization to the GFM's implicit 3D prior and generalization ability, which relies less on domain-specific appearance than 2D baselines. Real-world fine-tuning further improves performance, while \textbf{Ours-foundation} matches or exceeds \textbf{Ours-finetune}, suggesting that broader training can enable near out-of-the-box deployment. Training the closest baseline, VSCDNet, with the same strategy as Ours-foundation still yields substantially lower performance (19.27 vs. 62.54 change IoU on SceneDiff), supporting the importance of our architecture rather than training data alone.  Figure~\ref{fig:main_dataset_result} shows qualitative results.

\subsection{Video Scene Change Detection (VSCD)}
\customCDTable
Table~\ref{tab:custom_cd} shows that \modelName{} substantially outperforms all baselines in all metrics across all benchmark, while \textbf{Ours-foundation} retains a similarly strong advantage. Particularly, on \modelName{}-CD-SceneSmith, our model increase 42.01\% on change IoU and 27.91\% on F1.  These benchmarks are challenging due to large viewpoint changes and occlusions, where parallax and disocclusion can be mistaken for scene changes even after frame matching. By leveraging co-visibility-aware cross-session comparison and the GFM's implicit 3D prior, \modelName{} better distinguishes true scene changes from viewpoint-induced appearance differences. Qualitative results are shown in Appendix~\ref{app:additional_qualitative}.

\subsection{Scene Change Detection (SCD) for Paired-image}
\scdTable
Table~\ref{tab:scd} shows that, with only one image per session, \modelName{} achieves the highest change IoU and F1 on PSCD, ChangeSim, and VL-CMU-CD. These results indicate that \modelName{}'s advantage is not limited to additional video context: the GFM's implicit 3D features provide stronger change descriptors than conventional 2D features. The same architecture therefore delivers strong performance in both video and traditional paired-image change detection.

\subsection{Ablation Studies}
\ArchitectureAblationTable
Table~\ref{tab:architecture_ablation} ablates our architectural components. We first examine the benefit of GFM features. On VSCD, which contains large viewpoint differences, both GFM backbones, $\pi^3$ and VGGT-$\Omega$, substantially outperform the 2D backbones, SAM and DINOv2. On PSCD, although DINOv2 achieves performance comparable to $\pi^3$, VGGT-$\Omega$ still outperforms both 2D backbones. The relatively strong performance of DINOv2 on PSCD may be attributed to the limited viewpoint differences in this dataset. Overall, these results highlight the benefit of geometric pretraining, particularly under substantial viewpoint variations, which is not limited to a specific GFM backbone.

We further examine the contributions of individual architectural components. Even with a simple DPT Fusion Head, VGGT-$\Omega$ features outperform both 2D backbones, further highlighting the value of implicit 3D priors from geometric pretraining. On VSCD, Inter-session Cross-Attention further improves performance, providing an additional 7.49\% improvement in F1, with the Camera Reader providing an additional 3.13\% in F1 when combined with it, despite hurting performance in isolation. This suggests that pose information is most effective when used to guide cross-session comparison. The benefit is larger on video-based VSCD, where multi-view co-visibility helps resolve viewpoint and occlusion ambiguities. Additional analysis is provided in Appendix~\ref{app:additional_experiments}.

\subsection{Real-world Experiment}
\label{sec:real_world_exp}

\realworldFigure
To evaluate \modelName{}-SLAM in realistic settings, we conduct offline human-captured and online robotic experiments across small, medium, and multi-room indoor scenes with 4--10 object changes. Human trajectories use unposed iPhone videos, while robotic deployments use an Agilex mobile manipulator with an Intel RealSense D455. Both offline and online inference run on an NVIDIA RTX 5090 desktop GPU; during online deployment, the robot streams camera images to the workstation via ROS 2. As shown in Figure~\ref{fig:real_world_figure}, \modelName{}-SLAM operates under clutter, occlusion, and viewpoint changes, consistently localizing additions and removals across object scales in both image space and the reconstructed change-aware 4D map. With inputs capped at $64$ images, each change-detection forward pass takes only $0.59$\,s. During online deployment, the SLAM loop produces a new submap every $8.2$\,s on average, while the corresponding asynchronous change-detection job completes in $6.9$\,s, allowing change updates to keep pace with reconstruction and typically finish before the next submap is produced. The accompanying video demonstrates the full on-the-fly reconstruction and map-update process; additional results on other scene scales are provided in Appendix~\ref{app:real_world_exp}. Appendix~\ref{app:slam_quant} further evaluates \modelName{}-SLAM quantitatively on the \modelName{}-CD-SceneSmith test set, where it reaches $68.86$ change IoU and reduces ghost points by $38.2\%$ over naive fusion.

\section{Discussions}

\failureCaseFigure

\textbf{Illumination Changes.} To test robustness to lighting, we simulate illumination changes through data augmentation. Appendix~\ref{app:additional_failure} shows that the model is robust to mild changes but degrades under extreme ones. We also test the model on extreme real-world lighting changes. Figure~\ref{fig:failure_illumination} shows an outdoor scene captured at noon and at night, where VGGT-$\Omega$ fails to register the two visits, leading to poor change detection. This exposes a limitation of VGGT-$\Omega$ under extreme lighting.

\textbf{Semantic Changes.} In real-world trials, the model shows an emergent ability to detect some semantic changes, such as a laptop being turned on or off. However, Figure~\ref{fig:failure_semantic} shows that it is inconsistent on more semantic changes. We attribute this to our training data, which lacks such changes.

\textbf{Tiny Objects in Cluttered Scenes.} The model has difficulty detecting small objects in cluttered scenes. Figure~\ref{fig:failure_tiny} shows an example from our dataset in which a battery and a marker, marked by the red box, are removed. Both objects are small and far from the camera, and the viewpoint changes greatly between visits. The model detects the change only later in the sequence, when the camera moves closer to the marker. Our datasets do not specifically target such cases.

\textbf{Limitations and Future Directions.} Our work has three main limitations. First, an ideal change-aware 4D map should capture all changes, including dynamic objects, but we address only quasi-static changes between visits. Our datasets also mainly cover indoor, object-level geometric changes. Extending them to semantic changes (e.g., a light turned on or a bottle emptied), structural changes (e.g., repainted walls or new construction), and outdoor environments would broaden what the model can detect. Second, \modelName{} outputs binary masks without instance information, so it cannot detect moved objects. Grouping changes into instances would also enable compact representations such as scene graphs. Finally, \modelName{}-SLAM relies on SLAM to build a consistent map and retrieve reference frames. Removing this dependence would improve robustness to SLAM failures.

\section{Conclusion}
Detecting environmental changes and reconstructing consistent spatio-temporal maps remain critical yet challenging tasks for long-term robot autonomy. In this paper, we introduced \modelName{}, which leverages implicit 3D knowledge from GFMs and a co-visibility-aware spatio-temporal comparator for joint scene change detection and 3D reconstruction. Training across diverse datasets, including our new benchmark, enables strong generalization across learned and zero-shot settings and takes a step toward a foundation model for scene change detection. We further introduced \modelName{}-SLAM for online change-aware 4D mapping over long sequences, bridging change detection and long-term robot mapping. More broadly, our results demonstrate the potential of implicit geometric knowledge for change detection and encourage further research on other spatial reasoning tasks.

\subsubsection*{Acknowledgement}
This work was partially funded by ARL DCIST, ONR RAPID, and by VinMotion. We sincerely thank Nicholas Pfaff for providing support on SceneSmith, Dominic Maggio for providing support on VGGT-SLAM, and Paul Liang and Lukas Schmid for supporting ideation and feedback.

\bibliographystyle{iclr2027_conference}

\newpage
\appendix
\makeatletter
\setlength{\@fptop}{0pt}
\setlength{\@fpbot}{0pt plus 1fil}
\makeatother
\section{Appendix}

\subsection{Backbone Attention Comparison}
\label{app:backbone_attention}

\backboneAttentionFigure
Figure~\ref{fig:backbone_attention} qualitatively compares the backbones evaluated in Table~\ref{tab:architecture_ablation}. When querying the can, the geometric backbones (VGGT-$\Omega$ and $\pi^3$) show clean, sharp attention on the object in $v_1$ at certain layers. The 2D backbones (DINOv2 and SAM) also locate the can, but their attention is noisier and leaks into the background, making the localization ambiguous. The difference is clearer in $v_0$, where the can is absent and no correspondence should exist. Here, the geometric backbones consistently show low attention across layers. VGGT-$\Omega$ is especially consistent, while $\pi^3$ shows some weak, diffuse attention in the background at L13. In contrast, the 2D backbones still respond strongly: DINOv2 attends to the background and other objects at all layers, and SAM wrongly attends to other objects at L8. These observations support our hypothesis that geometric backbones are more effective at finding cross-visit correspondence. The cleaner attention of VGGT-$\Omega$ compared with $\pi^3$ may also explain the performance gap in the ablation, possibly because VGGT-$\Omega$ is trained with a matching loss that strengthens 3D correspondence.

\subsection{Model Implementation Details}
\label{app:model_implementation}

\textbf{Camera Reader.} Each $\mathrm{MLP}_{\ell}$ has two linear layers with a GELU activation between them, mapping width $2d$ to $d/2$ and back to $d$ ($4096\rightarrow1024\rightarrow2048$). Only the patch tokens are updated; the camera and register tokens of each layer pass through unchanged.

\textbf{Inter-session Cross-Attention.} Each $\mathrm{CrossAttn}_{\ell}$ uses $16$ heads ($d_h=d/16$) with pre-normalization, applying separate layer normalizations to the queries and to the keys/values before the learned projections. The tokens of all frames in a visit are flattened into a single sequence, so each token attends to every token of every frame in the other visit, and both directions read the original features $Z_{v_0}^{\ell}$ and $Z_{v_1}^{\ell}$. The feed-forward sublayer is
\begin{equation}
\tilde{Z}_{v}^{\ell} = \bar{Z}_{v}^{\ell} + \mathrm{FFN}_{\ell}\big(\mathrm{LN}_{\ell}(\bar{Z}_{v}^{\ell})\big), \qquad v\in\{v_0,v_1\},
\end{equation}
where $\mathrm{LN}_{\ell}$ denotes layer normalization and $\mathrm{FFN}_{\ell}$ is a two-layer network with GELU activation and hidden width $4d$.

\textbf{Loss Function.} The weighted binary cross-entropy is computed over the valid pixels of all images in a batch,
\begin{equation}
\mathcal{L} \;=\; \frac{1}{\sum_{r}|\Omega_r|}\sum_{r}\sum_{u \in \Omega_r} \mathrm{BCE}_{w^{+}}\!\big(z_r(u),\, M_r(u)\big),
\end{equation}
where $\Omega_r \subseteq \{1,\dots,H\}\times\{1,\dots,W\}$ is the set of valid pixels of image $r$, $M_r$ is its ground-truth mask, and the positive (change) class is up-weighted by $w^{+} = 5$. Frames without a pixel-aligned label (e.g., the unlabeled image of an SCD pair annotated in only one visit) receive $\Omega_r = \emptyset$ and are excluded from the loss.

\textbf{Training Details.} The DPT Fusion Head is warm-started from the pretrained depth head, whose DPT decoder has the same architecture, and only its final output layers are re-initialized; the Camera Reader and Inter-session Cross-Attention are randomly initialized. Each batch draws its number of images from $\{2, 8, 16, 32, 48, 64\}$ and an aspect ratio from $[0.5, 2.0]$, with the batch size chosen so that each GPU processes at most $64$ images at resolution $256$. The paired-image and video streams are served by separate loaders and interleaved by deficit scheduling: each step draws from the stream that has consumed the smaller fraction of its epoch. This spreads both data types evenly over training, giving a balanced learning signal throughout rather than long runs of a single type, and lets both streams finish together despite variable video batch sizes. For augmentation, we apply color jitter, a horizontal flip shared by all frames of a sample, and a random swap of the visit order with probability $0.5$, so the shared cross-attention weights see both directions equally. We use AdamW with weight decay $0.05$ and a peak learning rate of $10^{-5}$, with linear warm-up over the first $5\%$ of steps followed by cosine decay, gradient clipping at norm $1.0$, and bfloat16 mixed precision. We train for $50$ epochs with DDP on two RTX 4090 GPUs, which takes about three days.

\subsection{Dataset Generation}
\label{app:dataset_generation}
\datasetGenerationFigure

\textbf{Synthetic Dataset Generation.}
We build the synthetic portion of our dataset in five steps:
\begin{enumerate}
    \item \emph{State sampling.} For each scene, we generate three states by randomly removing movable objects. A support graph inferred from object geometry preserves physical plausibility by removing dependent objects together with their supports, while structural elements required for navigation are retained.
    
    \item \emph{Coverage planning.} Candidate viewpoints are sampled over the navigable space and evaluated for object visibility. We select viewpoints that jointly cover scene objects and navigable areas, then connect them with an open-path TSP to produce an independent trajectory for each state. The pipeline is illustrated in Figure~\ref{fig:dataset_generation}.
    
    \item \emph{Rendering.} Each state is rendered along its trajectory at $512\times512$ and $30$ fps, with camera orientation varying independently of travel direction to increase viewpoint diversity.
    
    \item \emph{Change masks.} Ground-truth masks are derived from object identities rather than pixel differences. For each ordered state pair, objects present in one state but absent from the other are marked as changed and rendered with a dedicated semantic label, yielding exact per-frame masks.
    
    \item \emph{Cross-visit matching.} Because visits follow independently generated trajectories, we match frames across states using scene overlap and pose similarity, retaining the best cross-visit correspondences for training and evaluation. Figure~\ref{fig:view_match} shows that matched views share substantial visible scene geometry.
\end{enumerate}

\datasetCovisibilityFigure

\modelName{}-CD-HSSD and \modelName{}-CD-SceneSmith share this generation pipeline, differing mainly in scene source and simulator. \modelName{}-CD-HSSD uses HSSD scenes~\citep{Khanna24cvpr-HSSD} rendered in Habitat~\citep{Savva19iccv-habitat, Szot21neurips-habitat}. \modelName{}-CD-SceneSmith applies the same pipeline to procedurally generated SceneSmith scenes~\citep{pfaff26icml-scenesmith}, using Drake's VTK renderer and a walkable navmesh derived from scene geometry. We provide both whole-house and per-room variants.

\textbf{Real-World Dataset Generation.}
We collect \modelName{}-CD-real from independently captured revisits of real indoor scenes. Changes are annotated using the SceneDiff tool~\citep{Wu25eccv-SceneDiff}: annotators identify changed objects and provide sparse point prompts, which SAM2 propagates across both videos to produce dense per-frame masks. This supports real-world additions, removals, modifications, and deformable changes. We modified the tool to make it more user-friendly to speed up annotation, so that others can collect and annotate more data in the same way. We refer readers to the original paper for further details on the annotation procedure.

\textbf{Dataset Statistics Comparison Protocol.}
Table~\ref{tab:dataset} compares our datasets with prior scene change detection benchmarks. \emph{Places} approximates the number of visually distinct locations visited within a trajectory. We cluster per-frame DINOv2--SALAD descriptors using a cosine-distance threshold of $0.7$ and merge revisits to the same location. \emph{Length} denotes the duration of a single visit, computed from its frame count at $30$ fps. Both are reported as mean/minimum/maximum over visits. \emph{Pairs} counts cross-visit comparisons: one pair corresponds to two images for SCD datasets or two video clips for VSCD datasets. \emph{Views} counts frames with ground-truth change masks.

\subsection{Online System Implementation Details}
\label{app:online_system}

We retain prior visit $v_0$ and reconstruct current visit $v_1$ online, submap-by-submap, using VGGT-SLAM~\citep{Maggio26rss-VGGT-SLAM2}. Each submap triggers a bounded \emph{change-detection (CD) job} that processes its keyframes with selected prior keyframes in one \modelName{} forward pass and transfers predictions to sparse 3D points.

\textbf{Cross-visit alignment.}
We reuse VGGT-SLAM's SALAD-based~\citep{Izquierdo24cvpr-salad} loop closure. Once a cross-visit closure aligns $v_1$ with the prior reconstruction from $v_0$, CD jobs can use geometrically comparable prior frames. Verified loop closures receive the highest pairing confidence, $c=2$.

\textbf{Coarse SALAD matching.}
Using cached SALAD descriptors, each current keyframe retrieves its nearest prior keyframe by $L_2$ distance $d$, accepting matches with $d<\delta=0.9$. Matching is bidirectional, and accepted pairs receive $c=\max(0,1-d/\delta)$ before geometric verification.

\textbf{Geometric reprojection gate.}
Candidate pairs are verified by unprojecting depth, reprojecting across views, and computing symmetric in-frame depth-consistent overlap. We reject overlap below $0.5$, use a $15\%$ depth tolerance, and assign accepted pairs their overlap score as $c\in[0,1]$. To ensure, fast compute, verification uses an $8\times8$ strided pixel grid ($\sim1/64$ of pixels); world points are computed once per frame and cached across comparisons.

\textbf{Scene-centroid coverage search.}
For unmatched frames, we index prior keyframes by the centroid of their visible world points and geometrically evaluate only the $k=10$ nearest centroids using the same reprojection score. This supports large viewpoint changes while reducing search from $\mathcal{O}(n_{v_0}n_{v_1})$ to $\mathcal{O}(n_{v_1}k)$.

\textbf{Temporal fallback.}
Remaining unmatched frames inherit the correspondence of the nearest matched frame within $w=5$ frames, with confidence
$c=0.1(1-\Delta/(w+1))<0.1$ for temporal distance $\Delta$. Frames without a correspondence are excluded from change inference and cross-visit fusion.

\textbf{Asynchronous CD execution.}
Matching and inference run on a background worker, keeping reconstruction responsive. Jobs respect \modelName{}'s $64$-image limit by reducing the current-visit window while keeping matched pairs together.

\textbf{Change-aware map fusion.}
Per-pixel change probabilities and pairing confidences are transferred to their corresponding sparse 3D points; for prior frames used in multiple jobs, we retain the highest confidence. After pose-graph optimization, observations within each world-space voxel are fused by a confidence-weighted mean. Pairing confidence measures correspondence quality and is distinct from the model's unused per-pixel confidence. For a queried visit $v$ with other visit $v'$, points from $v'$ whose fused probability exceeds a threshold $\tau$ form the changed set $\mathbb{P}^{v'}_{\Delta}$; removing it from the joint reconstruction yields the change-aware point-cloud map $\mathbb{P}(v)$ of Section~\ref{sec:online}.

\subsection{Evaluation Protocol}
\label{app:evaluation_protocol}

\textbf{Adapting SCD Baselines to VSCD Benchmarks.}
Paired-image SCD methods assume correspondence between their two inputs. To evaluate them on video benchmarks, we establish cross-visit pairs using SALAD~\citep{Izquierdo24cvpr-salad}: for each frame in one visit, we retrieve its best-matching frame from the other visit and apply the SCD model to the pair. We repeat this in both directions so that every frame in both visits receives a prediction.

\textbf{Benchmark Setup.}
For PSCD and VL-CMU-CD, we follow the per-image protocols of \citet{Lin25icra-RobustSCD}. For ChangeSim, we exclude the ``removed'' class from the ground-truth masks and report results under the original ChangeSim protocol in Table~\ref{tab:original_changesim}. For VSCD, we retain the original RGB videos but replace the original $v_1$-aligned labels with masks for both visits, where each mask marks content present in that visit and absent from the other. We uniformly sample $32$ frames per visit and report results under the original VSCD protocol separately in Table~\ref{tab:original_vscd}. For SceneDiff and \modelName{}-CD-real, we sample $16$ frames per visit. For the longer HSSD-CD and SceneSmith-CD sequences, we evaluate within fixed 48-frame temporal windows, sampling $16$ frames from the current visit and selecting corresponding prior-visit frames using the cached cross-visit matches.

\subsection{Additional experiments for \modelName{}}
\label{app:additional_experiments}

\oscdPrescreenTable
\oscdResultTable
\oscdComparison
\oscdFailureVis
\textbf{Comparison against O-SCD.}  O-SCD predicts both ``appeared'' and ``removed'' masks for mask-only query frames given a set of reference frames, a formulation that differs from ours in Section~\ref{sec:problem_formulation}; we therefore evaluate on the original VSCD dataset, which is consistent with O-SCD's output format. As shown in Table~\ref{tab:oscd_prescreen}, under the original protocol of 32 reference and 32 query images, O-SCD frequently fails to produce results due to COLMAP or query-frame camera-pose registration failures. We thus provide 128 reference frames---a more generous setting that favors O-SCD---and select the top-12 samples on which it consistently succeeds. For this evaluation, we train our model on VSCD to predict both ``appeared'' and ``removed'' masks, and Table~\ref{tab:oscd_result} shows that our method outperforms O-SCD even under settings that favor it. We further provide a qualitative comparison in Figure~\ref{fig:oscd_qualitative}.

We further analyze O-SCD qualitatively (Figure~\ref{fig:oscd_failure_vis}) and identify two limitations that frequently hurt its pipeline: (1) poor rendering quality produced by noisy camera-position estimates or by query frames that fall outside the distribution of the reference frames; and (2) incorrect pose registration caused by large scene changes, e.g., when most objects are removed, which yields a well-rendered but wrong image for change detection.

\originalChamgeSimTable
\vscdBaselineTable

\textbf{Additional benchmark results.}
We additionally evaluate \modelName{} under the original ChangeSim and VSCD protocols. For these experiments, \modelName{} predicts both ``appeared'' and ``removed'' regions, which are combined into a binary change mask. For VSCD, we follow the original evaluation protocol and additionally compare against two abandoned-object detection (AOD) methods, TCF-LMO~\citep{padilla23spic-TCF-LMO} and PBCD-MC~\citep{tavares25ieeeaccess-PBCD-MC}.

\modelName{} remains consistently stronger under these alternative output protocols. On ChangeSim (Table~\ref{tab:original_changesim}), it improves change IoU from 34.20 to 51.75 and F1 from 46.01 to 65.13 over the strongest baseline, TERDNet. On the original VSCD benchmark (Table~\ref{tab:original_vscd}), \modelName{} outperforms prior methods on both synthetic and real-world data and remains strong across variations in video length, graphics quality, and the number of changed objects. These results suggest that \modelName{} learns transferable change cues that remain effective across different annotation and output conventions.

\runtimeTable
\textbf{Runtime Comparison.} We measure mean wall-clock time (seconds) of one forward pass (batch size 1, single RTX 5090) as the total number of input images $N$ grows ($N/2$ reference and $N/2$ query frames); every method predicts a change mask for all $N$ frames. $^\dagger$ indicates zero-shot methods; SceneDiff's time includes loading its Pi3, SAM, and DINOv3 models due to multi-stage pipeline design and it runs out of host CPU memory (OOM, 125\,GB RAM) at $N{=}128$. $^\ddagger$ O-SCD reconstructs a 3D Gaussian scene from the reference session (via COLMAP structure-from-motion and 3DGS training) before online query localization, so its runtime is dominated by this offline step rather than the online inference itself; it fails (Fail) for $N{\leq}8$ because COLMAP cannot register a usable reconstruction from so few reference views.

\subsection{Quantitative Results for \modelName{}-SLAM}
\label{app:slam_quant}

\slamQuantTable
To evaluate \modelName{}-SLAM as a complete system, we run it on every \modelName{}-CD-SceneSmith test scene, following the same protocol as the real-world experiments in Section~\ref{sec:real_world_exp}.

\textbf{Quantitative Evaluation Method.} We evaluate on the $30$ \modelName{}-CD-SceneSmith test scenes ($26$ Room and $4$ House), each with $3$ state pairs, for $90$ pairs in total. Change detection is scored on every keyframe of both visits. Keyframes that the pairing cascade cannot match to the earlier visit are skipped and scored as unchanged, so missed retrievals lower the score. For map quality, we build a ground-truth point cloud for each scene state by back-projecting the ground-truth depth of its keyframes with their ground-truth poses. We then align the reconstructed map to the ground truth with a single Sim(3) transform shared by both visits, estimated from the keyframe camera centers and their ground-truth counterparts, so that any misalignment between visits remains visible in the metrics. Standard reconstruction metrics such as RMSE are dominated by SLAM pose and depth errors, and changed geometry covers only a small part of a large scene, so these metrics barely reflect the effect of change detection. We therefore count \emph{ghost points}, which directly measure whether the map reflects the correct scene state. A reconstructed point in the map of visit $v$ is a ghost point if it is farther than $5$\,cm from the ground-truth geometry of state $v$ but within $5$\,cm of the ground-truth geometry of the other state. In other words, it is stale geometry from the other visit rather than ordinary reconstruction error. We count ghost points in the maps of both visits. \emph{Naive} fuses both visits without removing any changes, and \emph{Removed} is the per-pair reduction in ghost points, averaged over all pairs.

\textbf{Change detection.} Table~\ref{tab:slam_cd} shows that \modelName{}-SLAM reaches $83.51$ mIoU and $59.05$ F1 overall, which is consistent with the benchmark results that evaluate each pair of video segments individually. This indicates that the full system supports correct frame retrieval and change detection. House scenes score lower because drift over their longer trajectories causes more keyframes to be wrongly skipped, as analyzed below.

\textbf{Map quality.} Fewer ghost points means that more changed geometry has been removed from the map. Table~\ref{tab:slam_ghost} shows that, compared with naively fusing both visits, \modelName{}-SLAM removes $38.2\%$ of ghost points overall. The reduction is smaller on House scenes because more of their keyframes are skipped, so fewer changes are detected directly.

\slamSkipRateFigure
\textbf{Effect of skipped keyframes.} Figure~\ref{fig:slam_skip_rate} groups the $90$ \modelName{}-CD-SceneSmith test pairs of Table~\ref{tab:slam_quant} by the fraction of their keyframes that the pairing cascade leaves without a prior-visit counterpart. We use five groups ($0$--$2\%$, $2$--$5\%$, $5$--$10\%$, $10$--$20\%$, and $\geq 20\%$) and pool the metrics within each group. Pairs with almost no skipped keyframes ($0$--$2\%$, $42$ Room pairs) reach $91.91$ mIoU and $68.69$ F1. Accuracy decreases as more keyframes are skipped, falling to $68.11$ mIoU and $38.57$ F1 for pairs skipping $20\%$ or more. Because skipped keyframes are scored as unchanged, every changed pixel they contain is a false negative, so change IoU and F1 fall faster than static IoU. This trend also reflects scene difficulty: all House pairs fall in the two most-skipped groups. The \modelName{}-CD-SceneSmith trajectories were designed for data generation rather than SLAM evaluation: the camera moves fast, consecutive frames are far apart, and each trajectory follows a single open path that never returns to earlier places (Figure~\ref{fig:dataset_generation}). VGGT-SLAM can therefore close loops only between nearby frames and cannot correct the drift that accumulates over the long House traversals. As expected, this drift causes many keyframes to be skipped and makes the remaining pairings less reliable. Such fast, sparse trajectories are uncommon in robot deployment, and skipped keyframes are much rarer on our real-world traversals, as shown below.

\slamRealWorldSkipTable
\textbf{Skipped keyframes in real-world deployment.} Table~\ref{tab:slam_realworld_skip} reports the skip rate on the real-world sequences of Section~\ref{sec:real_world_exp} and Appendix~\ref{app:real_world_exp}. These sequences have no ground-truth masks, so we report only how often keyframes are skipped, i.e., left without a prior-visit counterpart after the pairing cascade and therefore never assessed for change. The \emph{Both visits} column pools the keyframes of the two visits. Fewer than $2\%$ of keyframes are skipped in four of the five sequences. The full-floor loop, our largest sequence, which is processed at real-time rate, skips $6.2\%$, which corresponds to the $5$--$10\%$ group of Figure~\ref{fig:slam_skip_rate}.

\subsection{Additional Limitation Analysis}
\label{app:additional_failure}

\illuminationTable
\textbf{Quantitative Results for Illumination Changes.} We simulate illumination changes between visits by relighting every frame of the query session ($v_1$) as $I' = \mathrm{clip}(g\, I^{\gamma}, 0, 1)$, while keeping the reference session unchanged. We use four levels: mild dark $(g,\gamma)=(0.6, 1.2)$, extreme dark $(0.25, 1.6)$, mild bright $(1.1, 0.6)$, and extreme bright $(1.2, 0.4)$. Table~\ref{tab:illumination} shows the results of \textbf{Ours-foundation}. Performance is nearly unaffected under mild darkening and brightening, and degrades gracefully under extreme darkening. Extreme brightening causes the largest drop, reducing change IoU by $6.0\%$ on VSCD and $7.7\%$ on \modelName{}-CD-real relative to the unaltered input, because clipped highlights destroy image detail. However, these simulated changes are global, monotonic tone curves applied at test time. They preserve edges and relative contrast, and do not cover spatially varying real-world effects such as shadows, lights turned on or off, color casts, or low-light noise. Figure~\ref{fig:failure_illumination} in the main paper therefore also evaluates the model qualitatively under a real-world illumination change.

\homeExtremeFailureFigure
\textbf{Substantial Scene Rearrangement.} When a scene differs substantially from the prior visit, \modelName{}-SLAM struggles to establish correspondence and build an accurate 3D map. In Figure~\ref{fig:failure_home_extreme}, the \emph{Session 1 Only} and \emph{Session 2 Only} maps show the two scene states, between which the furniture was heavily rearranged. For example, both the desk and the bed moved to the opposite side of the room. On this scene, \modelName{}-SLAM drifts significantly in the left part of the map, which leads to noisy frame retrieval and in turn degrades change detection. It still produces sensible results in better-aligned areas, such as the kitchen in (1), and for the object changes in (3). However, it struggles to detect large changes, such as the rearranged table and bed.

\subsection{Additional Real-World Qualitative Results}
\label{app:real_world_exp}
Figure~\ref{fig:robot_workshop} and Figure~\ref{fig:31_altrium} show the mapping results from the Agilex mobile manipulator in small and medium environments. Figure~\ref{fig:human_office} and Figure~\ref{fig:human_home} show the offline mapping results from iPhone-captured data in a small office and a multi-room indoor apartment. 
\robotWorkshopFigure
\altriumFigure
\humanOfficeFigure
\humanHomeFigure

\subsection{Additional Qualitative Results From Change Detection Datasets}
\label{app:additional_qualitative}

We provide additional qualitative results of sampled scenes from each dataset. SCD results are shown in Figure~\ref{fig:scd_dataset_result}, and VSCD results are shown in Figure~\ref{fig:scenediff_result}, \ref{fig:vscd_dataset_result}, \ref{fig:argos_cd_real_result}, \ref{fig:hssd_cd_result}, and \ref{fig:scenesmith_result}.
\SCDResultFigure
\VSCDResultFigure

\end{document}